\documentclass[10pt,twocolumn,letterpaper]{article}

\usepackage{cvpr}      

\definecolor{cvprblue}{rgb}{0.21,0.49,0.74}
\usepackage[pagebackref,breaklinks,colorlinks,allcolors=cvprblue]{hyperref}

\usepackage{multirow}
\usepackage{booktabs}
\usepackage{tabularx}
\usepackage{makecell}
\usepackage{float}
\usepackage{hyperref}
\usepackage[accsupp]{axessibility}  

\def\paperID{*} 
\def\confName{CVPR}
\def\confYear{2026}

\title{AffordGen: Generating Diverse Demonstrations for Generalizable Object Manipulation with Affordance Correspondence}

\author{
Jiawei Zhang$^{1}$\thanks{Equal Contribution},
Kaizhe Hu$^{2,1}$\footnotemark[1],
Yingqian Huang$^{1,3}$,
Yuanchen Ju$^{4}$,
Zhengrong Xue$^{2,1}$,
Huazhe Xu$^{2,1}$\thanks{Corresponding Author} \\\\
$^1$Shanghai Qi Zhi Institute \quad
$^2$Tsinghua University \quad
$^3$Fudan University \quad
$^4$UC Berkeley
}
\definecolor{royalblue}{RGB}{65, 105, 225}
\definecolor{softgreen}{RGB}{85, 170, 85} 
\definecolor{softred}{RGB}{200, 50, 50}   

\hypersetup{
    colorlinks=true,
    linkcolor=orange,    
    citecolor=softgreen,    
    urlcolor=royalblue      
}

\newcommand{\ours}{AffordGen\xspace}

\begin{document}


\twocolumn[{%
\renewcommand\twocolumn[1][]{#1}%
\maketitle
\begin{center}
\includegraphics[width=0.97\linewidth]{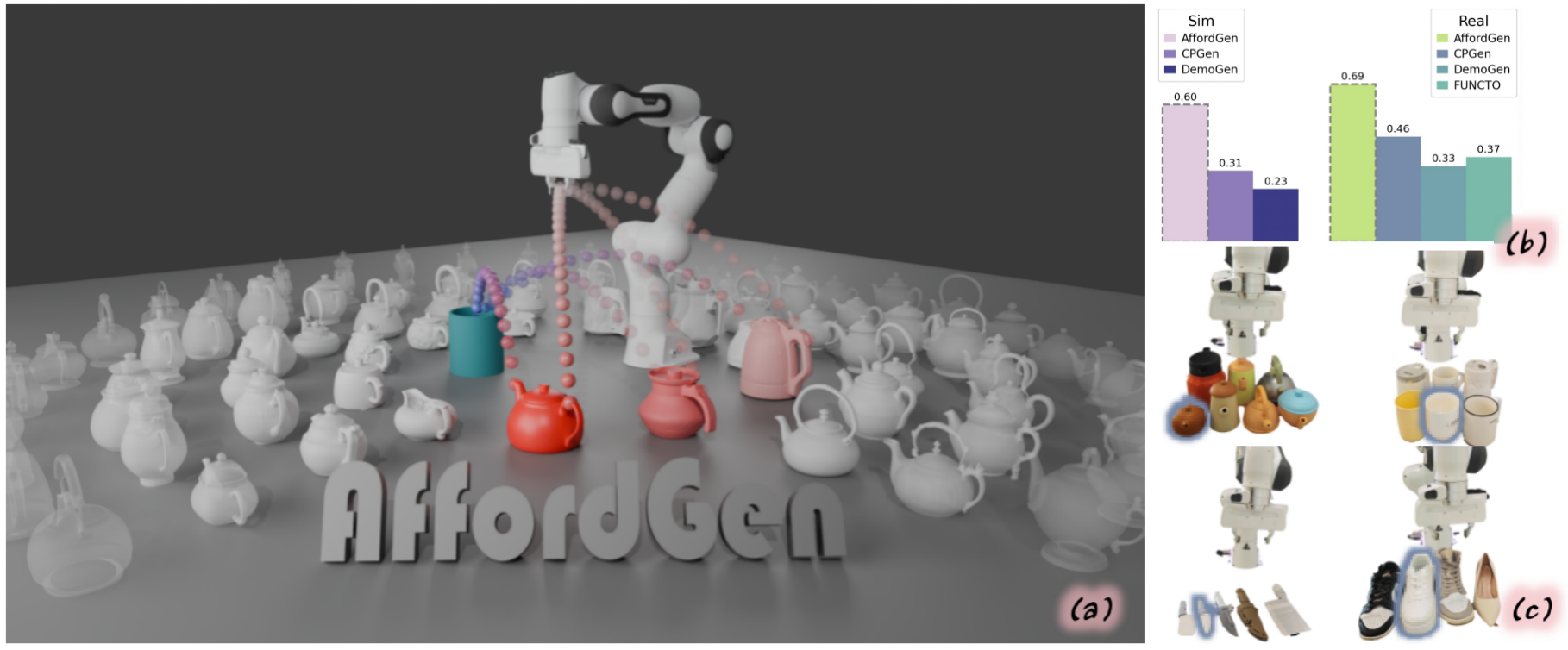}
\end{center}
\vspace{-0.5cm}
\captionof{figure}{\textbf{AffordGen overview.} (a) Diverse trajectory generation for novel objects via one-shot demonstration. (b) Superior performance against powerful baselines. (c) Real-world generalization to unseen objects from a single source.}
\label{fig:teaser}
\vspace{0.5cm}
}]
\begin{abstract}
Despite the recent success of modern imitation learning methods in robot manipulation, their performance is often constrained by geometric variations due to limited data diversity. Leveraging powerful 3D generative models and vision foundation models (VFMs), the proposed \textbf{AffordGen} framework overcomes this limitation by utilizing the semantic correspondence of meaningful keypoints across large-scale 3D meshes to generate new robot manipulation trajectories. This large-scale, affordance-aware dataset is then used to train a robust, closed-loop visuomotor policy, combining the semantic generalizability of affordances with the reactive robustness of end-to-end learning. Experiments in simulation and the real world show that policies trained with AffordGen achieve high success rates and enable zero-shot generalization to truly unseen objects, significantly improving data efficiency in robot learning.
\end{abstract}   
\maketitle
\section{Introduction}
Visuomotor imitation learning has achieved impressive progress in robotic manipulation~\citep{Chi2023DiffusionPV, ze20243d, Liu2024RDT1BAD, Wang2023MimicPlayLI, Lin2024DataSL, trilbmteam2025careful}. However, its practical deployment is hindered by two fundamental challenges: the prohibitive cost of collecting large-scale, high-quality human demonstrations, and the poor generalization of learned policies to novel objects and scenarios not encountered during training~\cite{Yuan2023RLViGenAR}. This reliance on extensive demonstration constrains the applicability of imitation learning in diverse real-world applications. 

Synthetic data generation offers one promising remedy. Recent works like DemoGen~\citep{Xue2025DemoGenSD} can expand a single demonstration into hundreds of spatially diverse trajectories, greatly improving data efficiency. Nevertheless, these approaches face a critical limitation: they primarily augment spatial relationships for a \textit{single} object instance. As a result, they inherit the restricted semantic scope of the source demonstration and generalize poorly beyond the source object. They also focus more on the translation invariance of the given task, exhibiting a limited ability to deal with different orientations.

Affordance-based methods, such as Robo-ABC~\citep{Ju2024RoboABCAG} and DenseMatcher~\citep{zhu2024densematcher}, leverage semantic correspondence to transfer affordance knowledge to unseen objects. They can transfer the manipulation method of one instance of an object to the other, enabling one-shot imitation learning methods with cross-category generalization, such as FUNCTO~\citep{Tang2025FUNCTOFO} possible. However, these approaches are typically planning-centric, and rely heavily on the accuracy of the mapped affordance point and the planning algorithm. The execution of the policy just follows the pre-computed, open-loop trajectories, lacking the reactive ability of learning-based closed-loop policies. Thus, while affordance research is highly active, a systematic approach to effectively integrate this semantic knowledge into learning-based pipelines is still missing.

In this work, we introduce \ours, a novel framework that repurposes affordance information as a generative prior for policy learning. \ours produces feasible training data on unseen objects that may come from a different category. Starting from a few human demonstrations, we establish correspondences between the demonstrated object’s keypoints and a large set of unseen 3D models. Then, we synthesize diverse yet semantically grounded trajectories in the style of DemoGen, but crucially over novel object instances and full 6D spatial relations. In this way, \ours scales a handful of demonstrations into thousands of high-quality, affordance-aware trajectories that cover a diverse realm of objects.

The key innovation of \ours lies in its novel use of affordance knowledge to generate trajectories. We argue that affordance knowledge is best utilized not through online planning, but as a powerful guide of data generation. By leveraging affordances to generate a large, diverse dataset of plausible trajectories, we can then train a reactive, closed-loop visuomotor policy that inherits both the semantic generalizability of affordances and the robustness of end-to-end learning. 

\ours demonstrates impressive performance on both simulation and real-world tasks, achieving an average performance boost of \textbf{24.1\%} and \textbf{24.3\%} over the best baseline on simulation and real tasks. It can generate thousands of meaningful trajectories over hundreds of different meshes, achieving a high success rate from as few as one source demonstration. AffordGen boosts the model performance on both generated ``seen'' objects and truly unseen objects that are not in the generated dataset. Beyond these abilities, it also has the potential to generate new data on completely different objects, as long as they share the same manipulation type. We also find that the object-level generation ability increases first and then decreases as we extend the generation to more unseen objects, guiding future cross-instance generation works. The contribution of \ours can be summarized below: 
\begin{itemize}[leftmargin=8mm]
    \item We introduce AffordGen, a novel framework that repurposes affordance correspondence as a generative source, enabling the synthesis of diverse and semantically meaningful demonstrations.
    
    \item We demonstrate that our method can scale a minimal number of human demonstrations into thousands of trajectories across novel object categories and full 6D poses, overcoming the semantic and geometric limitations of prior data augmentation techniques.
    
    \item We validate through extensive experiments that policies trained with \ours achieve significant zero-shot generalization to unseen objects, presenting a new paradigm for data-efficient robot learning.
   
\end{itemize}  
\section{Related Works}
\label{sec:related}
\subsection{Data Generation for Robot Manipulation}

Automated data generation techniques for robot manipulation have great potential to alleviate the data problem of embodied AI. These methods primarily fall into two categories: adapting existing demonstrations and generating new data from scratch. 

The first paradigm, trajectory adaptation, is exemplified by MimicGen~\cite{mandlekar2023mimicgen} and its variants~\cite{jiang2025dexmimicgen, garrett2024skillmimicgen, nasiriany2024robocasa}, which segment a few source demonstrations into object-centric skills and replay them in new spatial configurations. While powerful, these methods rely on costly on-robot or in-simulation rollouts to capture the final interaction data. DemoGen~\cite{Xue2025DemoGenSD} addresses this bottleneck by introducing a fully synthetic pipeline that generates paired actions and observations from direct 3D point cloud editing. More related to our approach, CPGen~\cite{lin2025constraint} extends DemoGen by stretching and transforming the source mesh to improve the diversity of generated data and the generalization ability of the policy. These methods, however, are still limited to the source object and have poor generalization ability to unseen objects even within the same category. 

The second paradigm leverages large-scale generative models to create data from scratch. Systems like GenSim~\cite{wang2024gensim}, GenSim2~\cite{hua2024gensim2}, and RoboGen~\cite{wang2024robogen} use Large Language Models (LLMs) to propose novel tasks, generate corresponding scenes, and script automated solvers to create demonstrations. This approach excels at generating task-level diversity but is limited by the capability of the underlying automated solvers, which may not match the quality of human demonstrations. Orthogonal to this, another line of work~\cite{chen2023genaug, yu2023scaling, mandi2023cacti, hu2024generalizable} uses diffusion models to augment the visual appearance of existing data, enhancing a policy's visual robustness but not its ability to generalize to new spatial configurations or object types.

\begin{figure*}[ht]
\begin{center}
\includegraphics[width=15cm]{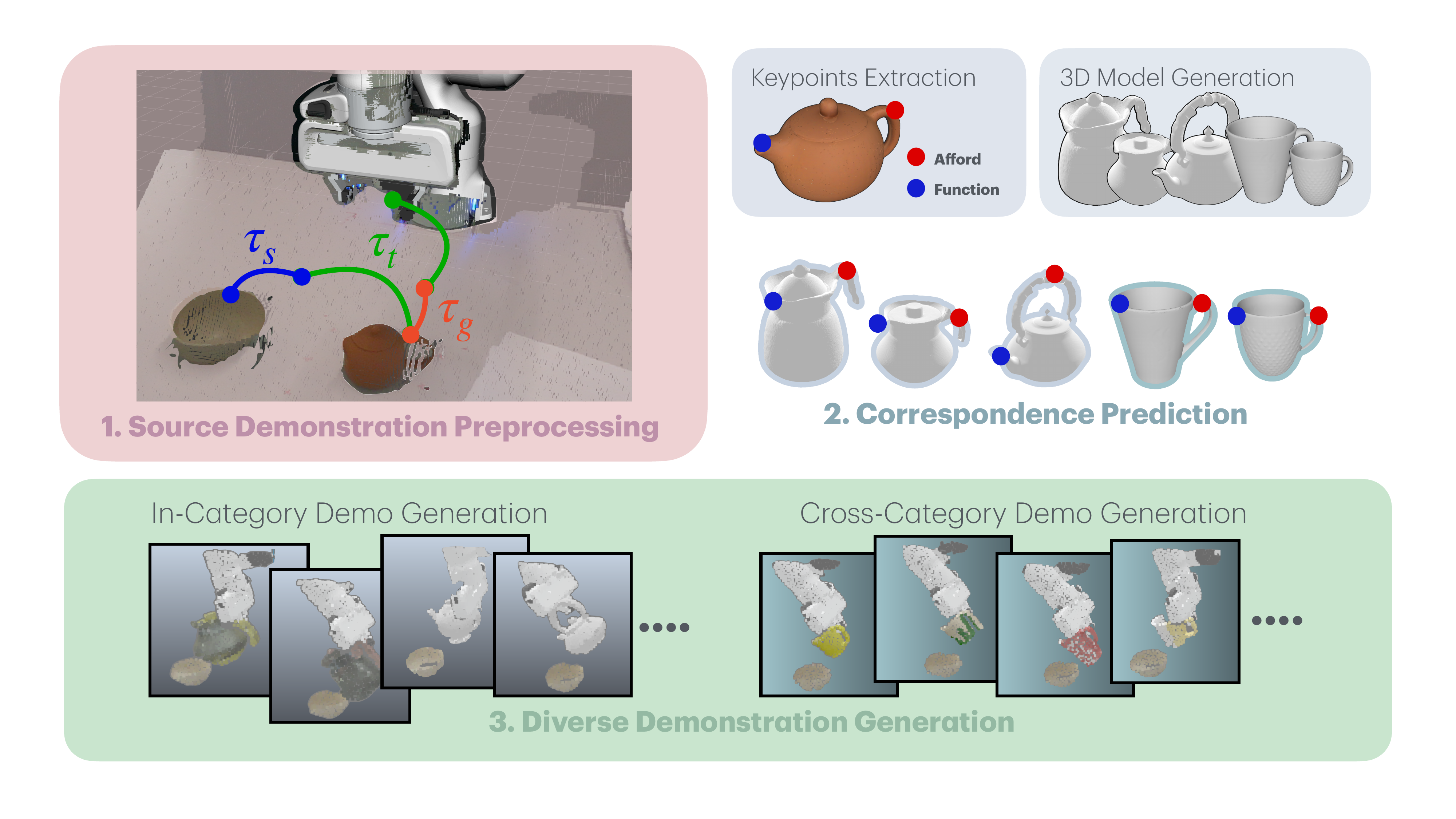}
\end{center}
\caption{1. \ours takes in a source expert demonstration and splits it into different functioning segments. 2. We extract keypoints on the source object and establish correspondences between them and many target objects in 3D space through visual foundation models. 3. By transferring the task-related segments and planning the transition segments, \ours can generate diverse and large-scale trajectories on both in-category and cross-category objects. }
\end{figure*}

\subsection{Semantic Correspondence for Manipulation}

Early research on semantic correspondence for manipulation relies on visual descriptors and semantic keypoints, such as DON~\citep{florence2018dense}, kPAM~\citep{manuelli2019kpam}, and NDF~\citep{simeonov2022neural}, which enabled instance-level transfer by aligning consistent geometric or appearance features. Current works exploit foundation models like diffusion models~\citep{tang2023emergent,zhang2023tale} for semantic alignment, and recent studies have extended such capabilities to robotic manipulation by leveraging semantic correspondence for affordance transfer across objects and categories~\citep{Ju2024RoboABCAG,zhu2024densematcher,kuang2024ram,wu2025afforddp,Tang2025FUNCTOFO}. Beyond these, other lines of research further explore semantic correspondence in different contexts: MimicFunc~\citep{tang2025mimicfunc} transfers functional correspondences from human videos to novel tools, and HRP leverages human affordances for robotic pre-training~\citep{srirama2024hrp}. 

Although these methods expand the scope of semantic correspondence, most still treat affordances merely as mapping signals, using them to locate contact points and relying on planners for execution, thus lacking true reactivity and flexibility. In contrast, \ours transforms affordances from static mappings into generative sources, synthesizing large-scale, diverse, and semantically consistent demonstrations that not only span across instances and categories but also provide rich data for training visuomotor policies, thereby achieving stronger generalization and adaptability.
\section{Methodology}
\label{sec:method}

\subsection{Problem Formulation}

Like many prior online planning works \citep{Tang2025FUNCTOFO, pan2025omnimanip} for robotic manipulation, we decompose a manipulation task into three distinct stages $\Omega = \{ \Omega_{G}, \Omega_{S}, \Omega_{T} \}$, where $\Omega_{G}$ denotes the grasp stage during which the robot closes its gripper to secure the object. $\Omega_{S}$ denotes the skill stage where the robot manipulates the grasped object to accomplish the task, like pouring tea into the tea cup, and $\Omega_{T}$ denotes the transition stage that connects the two other stages without collision. At each timestep $t$ within a stage, the robot takes in its current visual observation $o^{e}_{t}$ and proprioception observation $o^{s}_{t}$, and outputs a corresponding action $a_{t}$. AffordGen uses point clouds as the input type of $o^{e}_{t}$ because of its simplicity and special structure in 3D space that can be easily manipulated to generate new data.

The generation strategy of AffordGen is inspired by the intuition shared across the grasp and skill stages: despite the shapes and sizes of the objects manipulated in a given task varying, the semantic information embedded in their trajectories remains similar. In the following sections, we will discuss: (1) how to extract semantic information from the grasp and skill stages in the original demonstrations; (2) how to map this information onto large-scale 3D meshes different from the source; and (3) how to generate a large set of diverse demonstrations from the source trajectory.

\subsection{Source Demonstration Pre-processing}

Given an expert demonstration, we consider extracting three types of information: the \emph{grasping time} $t_{\text{grasp}}$ at which the gripper closes during $\Omega_{G}$; the \emph{skill segment} $\tau_{s}$ where the actions crucial to the task's success are performed; and the keypoints of the manipulated object in the original data, namely the \emph{affording point} and the \emph{function point}. 

The grasping time $t_{\text{grasp}}$ can be directly extracted according to the end-effector state information from the expert demonstration. The skill segment $\tau_{s}$ can be detected either through the video reasoning capabilities of vision-language models (VLMs) or by direct human annotation. The \emph{affording point} refers to the contact point between the gripper and the object, while the \emph{function point} refers to the point the tool interacts with other objects to accomplish the task. Both points are defined in 3D space, and can be readily recognized and annotated by VLMs or humans. As we only have a limited number of demonstrations, this process remains straightforward and efficient.

The point cloud used as visual input is derived from an RGB-D camera. On the raw RGB images, we employ SAM2 \citep{ravi2024sam2} to segment the objects involved in the task into four categories: robot, object, goal, and others. The segmentation labels on the 2D images are then mapped onto the point cloud along with the pixel coordinates, resulting in a segmented point cloud. Following the settings of DP3~\cite{ze20243d} and DemoGen, we further remove the background and floor from the obtained point cloud and then use Farthest Point Sampling (FPS) to down-sample it, thereby extracting the workspace point cloud $\mathcal{O}^{e} \subset \mathbb{R}^3$, which contains only the points located within the robot’s workspace.

\subsection{Semantic Correspondence on 3D Meshes}

\begin{figure}[ht]
\begin{center}
\includegraphics[width=8cm]{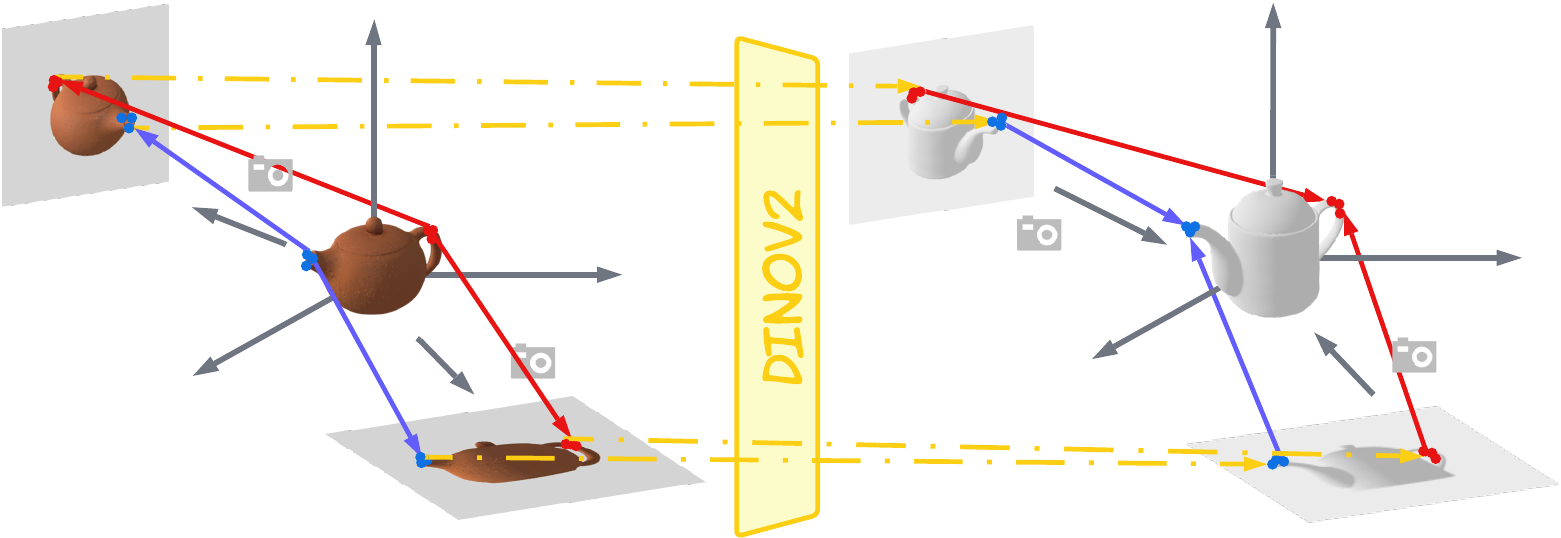}
\end{center}
\caption{Keypoints Correspondence in 3D Canonical Space. The keypoints are mapped to the target mesh through DINOv2.}
\end{figure}

To enable the reuse of trajectory information from the source demonstration, we need to establish correspondences between the keypoints of the manipulated object in the source demo and those on the new 3D meshes. While semantic correspondence across 2D images has been extensively studied, robotic manipulation tasks require accurate annotations in 3D space. Recent works have begun to explore semantic correspondences across 3D meshes~\cite{zhu2024densematcher}, but these approaches have been trained only on small-scale datasets, limiting their applicability to precise robotic manipulation tasks. To address this limitation, we propose a simple yet effective approach that normalizes all 3D meshes into a unified canonical space before mapping the 2D correspondences onto the 3D meshes.

Formally, let the source mesh be denoted as $M_{\text{src}}$, and let its associated keypoint be 
$x \in \mathbb{R}^3$, which lies in the local frame of $M_{\text{src}}$. 
Our objective is to find the corresponding keypoint $x'$ on the target mesh $M_{\text{tg}}$. For any given mesh, its pose is first normalized into the canonical space using a 6D pose estimator \citep{zhang2024omni6dpose}. We then perform parallel rendering to obtain RGB-D images from $n$ different camera views. 
The pose of the $i$-th camera is denoted as $P_i$ $(i = 1, 2, \dots, n)$, and the corresponding rendered image as $I_i$. Each image $I_i$ is fed into DINOv2~\cite{oquab2024dinov2} to obtain its semantic representation $S_i$. We select $m$ nearest mesh vertices in the neighborhood of $x$, denoted as 
$v_j \ (j = 1, \dots, m)$. Each vertex $v_j$ is projected onto the image 
$I_i$ via forward camera projection, resulting in the pixel coordinate 
$u_{ij}$. The corresponding pixel of $v_j$ on the target image is obtained by maximizing 
the cosine similarity in the feature space of the DINOv2 model:

\begin{align*}
u^{\text{tg}}_{ij} &= \arg\max_{u} \ 
\operatorname{CosSim}\big( S_i^{src}[u_{ij}], \ S_i^{tg}[u] \big), \\
w_{ij} &= \operatorname{CosSim}\big( S_i^{src}[u_{ij}], \ S_i^{tg}[u^{\text{tg}}_{ij}] \big).
\end{align*}

All the matched pixels $u^{\text{tg}}_{ij}$ are then unprojected into the 
3D space, yielding candidate correspondences $v^{\text{tg}}_{ij}$ with 
associated similarity scores $w_{ij}$. Finally, the target keypoint $x'$ in 3D space
is obtained by weighted average:
\[
x' = \frac{\sum_{i,j} w_{ij} \, v^{\text{tg}}_{ij}}{\sum_{i,j} w_{ij}}.
\]

\subsection{Diverse Demonstration Generation}

We follow a three-step approach to generate demonstrations on new meshes. We consider a robotic manipulation scenario where the end-effector manipulates an \textit{active object} to interact with a \textit{goal object}.
In this framework, the \textit{active object} is defined as the object being directly maneuvered by the robot (e.g., a teapot), while the \textit{goal object} is the object that the active object acts upon (e.g., a cup).

\subsubsection{Keypoint-constrained Trajectory Replay}

\begin{figure}[ht]
\begin{center}
\includegraphics[width=8cm]{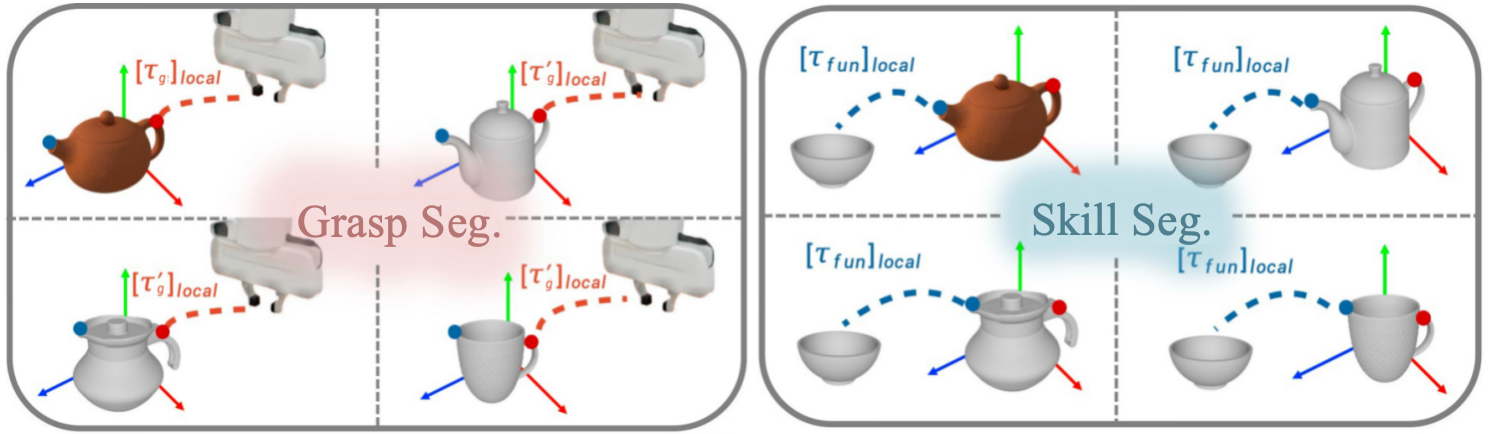}
\end{center}
\caption{Trajectory replay for grasp and skill segments. $\tau_g$ and $\tau_s$ remain similar across all meshes.}
\label{fig:traj_plan}
\end{figure}

Our goal is to efficiently transfer the grasp segment $\tau_g$ and the skill segment $\tau_s$ from source demonstrations to novel objects. 
We leverage the correspondence between the affording points ($x_{\text{aff}}$, $x_{\text{aff}}'$) and the function points ($x_{\text{fun}}$, $x_{\text{fun}}'$), following a simple yet effective assumption: 
when manipulating objects of the same \textit{function class}, the trajectories of the end effector relative to the affording point remain similar, while the trajectories of the function point relative to the goal object remain similar as well.
The \textit{function class} here extends the definition of object categories and refers to objects that share similar functionality. For example, a mug and a teapot share the same function of pouring water into a cup. An illustration is shown in Figure \ref{fig:traj_plan}.

To transfer \textbf{the grasp segment} $\tau_{g}$, we first normalize the end-effector segment $\tau_g$ into the local frame of the source mesh by using the initial pose of the source object $T_\text{init}$. The normalized grasp segment can be expressed as $\left [ \tau_g \right ] _\text{local} = T_\text{init}^{-1} \cdot \left [\tau_g \right ]_\text{world}$. Given a new object mesh $M_\text{tg}$ with the associated affording point $x_\text{aff}'$, the new grasp segment in the local frame of $M_\text{tg}$ can be obtained by
\[
\left [ \tau_g' \right ] _\text{local} = \left [ \tau_g \right ] _\text{local} - x_\text{aff} + x_\text{aff}'
\]

To transfer \textbf{the skill segment} $\tau_{s}$, we first derive the trajectory of the function point during the skill stage $\Omega_{S}$ in the world frame by leveraging its relative transformation to the end effector $T^{\text{fun}}_\text{ee}$. Thus, the function point trajectory in both world and local frame can be expressed as
\begin{align*}
    [\tau_{\text{fun}}]_{\text{world}} &= T^{\text{fun}}_{\text{ee}} \cdot [\tau_s] \\
    [\tau_{\text{fun}}]_{\text{local}}  &= T_{\text{init}}^{-1} \cdot [\tau_{\text{fun}}]_{\text{world}}
\end{align*}
respectively. For the new mesh $M_\text{tg}$ and its function point $x_\text{fun}'$, the new function trajectory in the local frame of $M_\text{tg}$ can be obtained by
\[
\left [ \tau_\text{fun}' \right ] _\text{local} = \left [ \tau_\text{fun} \right ] _\text{local} - x_\text{fun} + x_\text{fun}'
\]

For any random pose configuration $T'$ of mesh $M_\text{tg}$, the new grasp segment and skill segment can be solved from $\left [ \tau_g' \right ] _\text{local}$ and $\left [ \tau_\text{fun}' \right ] _\text{local}$ by

\begin{align*}
    [\tau_g'] &= T' \cdot \left [ \tau_g' \right ] _\text{local}\\
    [\tau_s'] &= T_\text{fun}^{\text{ee}} \cdot T' \cdot \left [ \tau_\text{fun} ' \right ] _\text{local}
\end{align*}

The resulting $\tau_g'$ and $\tau_s'$ correspond to the sequence of end-effector poses in world frame for executing the skill on the new mesh. The joint positions for each waypoint can be further computed via an inverse kinematics (IK) solver.

\subsubsection{Motion Planning for Transition Segment}
In most robot manipulation tasks, the connection segment between the grasp segment and the skill segment contains much less meaningful information. Such trajectories do not involve dynamic interactions between the robot arm and the active object, nor between the active object and the goal object. Therefore, it can be regarded as a collision-free free-space motion between the grasping segment and the skill segment. We employ motion planning to plan this trajectory or utilize spherical linear interpolation~\cite{shoemake1985slerp} to directly interpolate the path. We denote the resulting trajectory as $\tau_{m_i}$, where $i$ represents the $i$-th free-motion segment during the task. For example, the free motion segment between $\tau_g'$ and $\tau_s'$ can be expressed as 
\[
\tau_m' = \operatorname{MotionPlan} \left (\tau_g'[-1], \tau_s'[0] \right)
\]
\subsubsection{Point Cloud Digital Cousin Generation} 

\begin{figure*}[ht]
\begin{center}
\includegraphics[width=\linewidth]{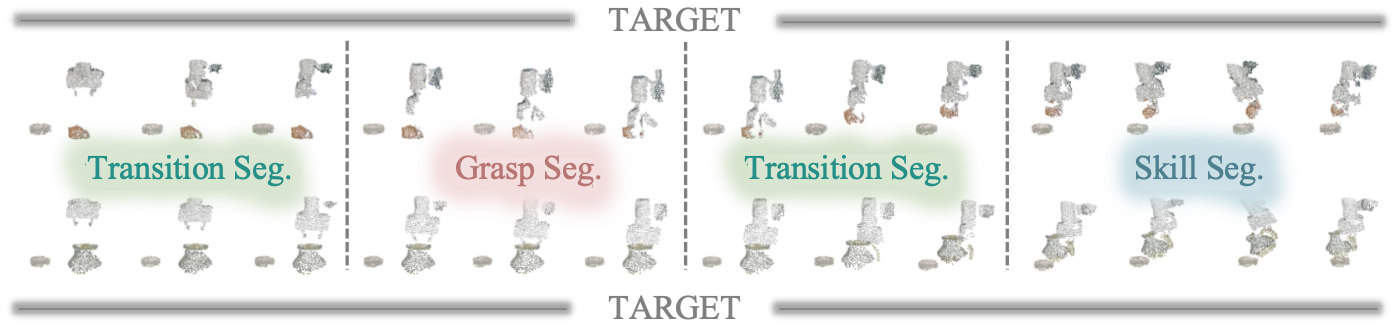}
\end{center}
\caption{Visualization of source and generated trajectory of the teapot pouring task. The upper line is the source trajectory, while the lower line is a generated trajectory. Each trajectory is composed of three types of segments.}
\label{fig:demo_vis}
\end{figure*}

After obtaining trajectories $\tau$ for novel objects, it is necessary to transform the point cloud of the source data accordingly to align with these new trajectories and meshes. DemoGen\cite{Xue2025DemoGenSD} achieves the generation of point clouds for new trajectories by applying simple global translation and rotation on the original point cloud. This approach is effective because the trajectories generated by DemoGen differ from the original ones primarily by a positional shift. In contrast, AffordGen aims to produce a variety of 3D models with diverse 6D poses within the workspace. To achieve this, we directly render the robot and the manipulated object point clouds from simulation, and then replace their counterparts in the source demonstration. The resulting hybrid real-simulated point cloud data mitigates the sim-to-real gap while bypassing the tedious process of fully reconstructing the environment in simulation. We employ parallel rendering to generate point clouds to achieve high throughput efficiency. A visualization of the generated trajectory, in comparison to the source demonstration, is provided in Figure~\ref{fig:demo_vis}.

Apart from these approaches, we apply special modifications of the skill segment to deal with the occlusion problem. For a detailed discussion of the data generation scheme and the generation quality, please refer to Appendix~\ref{app:data_generation}

\section{Experiments}

To validate our claims about the effectiveness of AffordGen in generating data through the transfer of grasp segments and skill segments, we conduct generalization experiments in both simulation and the real world.  

\subsection{Simulation Experiments}  

\subsubsection{Experiment Settings}

    
    

\begin{figure}[ht]
\begin{center}
\includegraphics[width=8cm]{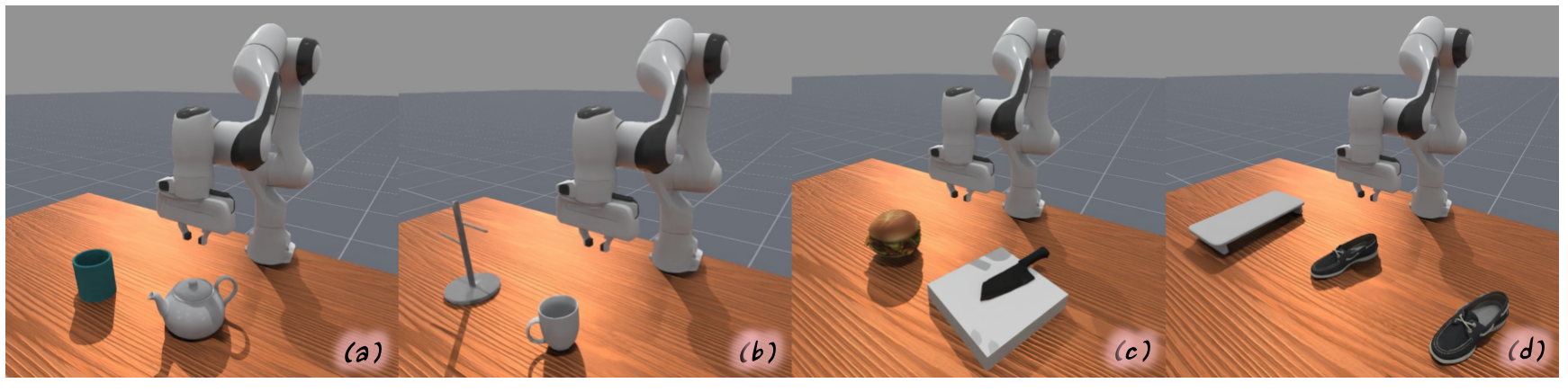}
\end{center}
\caption{Simulative experiments setup: (a) Teapot Pouring, (b) Mug Hanging, (c) Knife Cutting, (d) Shoe Organizing.}
\label{fig:sim_tasks}
\end{figure}

 We use ManiSkill3 \citep{taomaniskill3} as the simulator and construct four tasks involving both grasp and skill stages: \textit{teapot pouring}, \textit{mug hanging}, \textit{knife cutting} and \textit{shoe organizing}, as illustrated in Figure \ref{fig:sim_tasks}. For \textit{teapot pouring}, the robot needs to grasp the handle part of the teapot and then move and tilt the teapot to position its spout above the target cup. For \textit{mug hanging}, the robot needs to grasp the mug by its handle, move it, and insert the rack prong into the hole of the mug handle. For \textit{knife cutting}, the robot first grasps the handle of the knife and then maneuvers it to make contact between the blade and the target food. For \textit{shoe organizing}, which is a long-horizon and multi-object task, the robot grasps the shoes at the heels and places both the nearer and the farther shoes onto the shoe rack in correct orientation.
 
 For each task, we collect \textbf{one} expert demonstration as a reference for generation and generate \textbf{1000} demonstrations using 3D mesh assets, from the PAM \citep{zhang2024omni6dpose} dataset or 3D generative model \citep{hunyuan3d22025tencent}. The 3D meshes are randomly split into \textit{seen} and \textit{unseen} subsets; the former is used for data generation and the latter for evaluation. For detailed information on the 3D mesh dataset split and visualization, please refer to Appendix~\ref{app:dataset}.

 We compare the performance with DemoGen \cite{Xue2025DemoGenSD} and CPGen \cite{lin2025constraint}. Similar to AffordGen, DemoGen also operates on 3D point clouds. We use the same source demonstration and generate 1000 trajectories using DemoGen. 
 We take DemoGen as one of the baselines to demonstrate that AffordGen maintains its spatial generalization capability on the original mesh when generalizing to unseen shapes. CPGen generates demonstrations by stretching and transforming the original object mesh, aiming to enhance shape generalization. We also generate 1000 trajectories, but stretch them randomly in the scale range of the evaluation set. Note that the original CPGen operates on RGBD data. For a fair comparison, we adapt CPGen to work with the 3D point cloud modality. The implementation details of the baselines can be found in Appendix~\ref{app:exp_details}.

\subsubsection{In-category Generalization Results}

\begin{table*}[htbp]
\centering
\renewcommand{\arraystretch}{1.3}
\setlength{\tabcolsep}{6pt} 

\resizebox{\textwidth}{!}{
\begin{tabular}{c cc cc cc cc}
\toprule
\multirow{2}{*}{\#Mesh $\times$ \#Demo} 
& \multicolumn{2}{c}{\textbf{Teapot Pouring}} 
& \multicolumn{2}{c}{\textbf{Mug Hanging}} 
& \multicolumn{2}{c}{\textbf{Knife Cutting}} 
& \multicolumn{2}{c}{\textbf{Shoe Aligning}} \\
\cmidrule(lr){2-3}
\cmidrule(lr){4-5}
\cmidrule(lr){6-7}
\cmidrule(lr){8-9}
& Source & Unseen
& Source & Unseen
& Source & Unseen
& Source & Unseen \\
\midrule
DemoGen (1$\times$1000) & $0.933 \pm 0.009 $ & $0.131 \pm 0.029$ & $0.940 \pm 0.043$ & $0.402 \pm 0.036$ & $0.490 \pm 0.037$ & $0.224 \pm 0.012$  & $0.400\pm0.050$ & $0.212\pm0.025$\\
CPGen (1000$\times$1) & $0.713 \pm 0.146 $ & $0.169 \pm 0.070$ & $0.900 \pm 0.056$ & $0.502 \pm 0.027$ & $0.747 \pm 0.148$ & $ 0.424\pm 0.003 $ & $0.550\pm0.007$ &  $0.266\pm0.024$  \\
AffordGen (20$\times$50) & $0.920 \pm 0.086$ & $0.353 \pm 0.103$ & $\textbf{0.967} \pm 0.025$ & $0.683 \pm 0.004$ & $\textbf{0.793}\pm0.057$ & $\textbf{0.565}\pm0.002$ &$0.550\pm0.100$ &$0.438\pm0.013$ \\
AffordGen (50$\times$20) & $0.927 \pm 0.025$ & $\textbf{0.553} \pm 0.039$ & $0.777 \pm 0.021$ & $0.664 \pm 0.033$ & $0.647\pm0.068$ & $0.535\pm0.006$ & $0.588 \pm 0.018 $ & $0.302 \pm 0.089$\\
AffordGen (100$\times$10) & $\textbf{0.960} \pm 0.043$ & $0.519 \pm 0.072$ & $0.753 \pm 0.098$ & $\textbf{0.707} \pm 0.011$ & $0.606\pm0.090$ & $0.510\pm0.001$ & $\textbf{0.825}\pm0.035$  &$\textbf{0.588}\pm0.018$ \\
AffordGen (1000$\times$1) & $0.700 \pm 0.123$ & $0.242 \pm 0.067$ & $0.853 \pm 0.082$ & $0.642 \pm 0.018$ & $0.607\pm0.207$ & $0.542\pm0.002$ &$0.625\pm 0.025$ &$0.425\pm0.175$ \\
\bottomrule
\end{tabular}
} 

\caption{Comparison of different Mesh-Demo configurations across simulative tasks.}
\label{tab:sim_incat}
\end{table*}

To thoroughly examine the efficiency of \ours on the generalization capability of the learned policy, we perform ablation studies on each task by fixing the total number of demonstrations while varying the number of meshes used for generation and the number of demonstrations per mesh. Under each \textbf{$(\text{\#meshes}, \text{\#demos per mesh})$} combination, the learned policy is evaluated on both source mesh and 5-20 unseen meshes, where each mesh is tested 50 times with randomly placed initial positions and orientations.

The detailed performance is shown in Table \ref{tab:sim_incat}. We separate the results into two categories: ``source'' and ``unseen'', corresponding to the performance on the source mesh and the evaluation set of meshes. DemoGen, CPGen, and \ours all achieve spatial generalization on the source mesh, preserving the task information of the source object over the entire workspace. Despite being trained on extensive data for the source mesh, \ours achieves comparable performance to DemoGen. By applying stretching and compression transformations, CPGen augments the shape of the original object, leading to better performance than DemoGen on unseen object tests. AffordGen, through semantic-corresponding keypoints, significantly expands the shape diversity of the generated data, outperforming both DemoGen and CPGen across all tasks by an average of $\textbf{24.1\%}$  on unseen object tests on the best $100\times 10$ setting.

\begin{figure}[ht]
\begin{center}
\includegraphics[width=8cm]{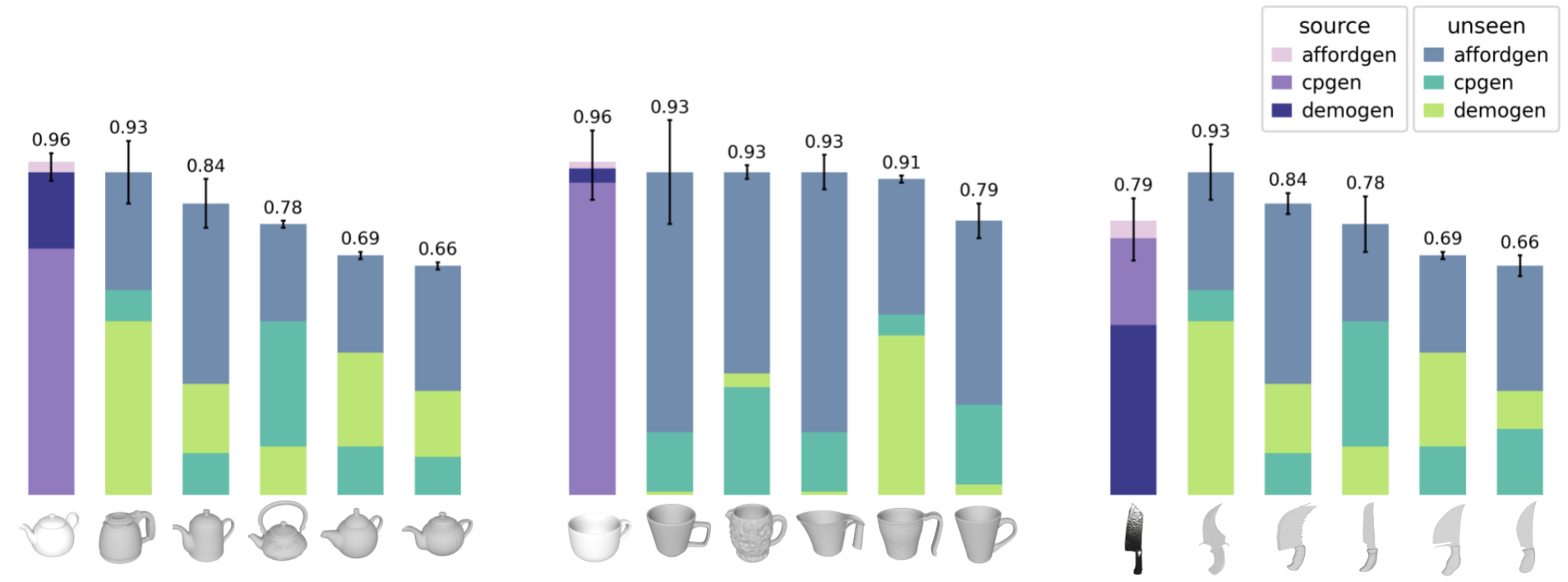}
\end{center}
\caption{Simulative evaluation results on different meshes}
\label{fig:results_per_mesh}
\end{figure}

To further demonstrate the generalization range of our method, we select 5 evaluation meshes and arrange them based on their performance in Figure~\ref{fig:results_per_mesh}. The performance gradually drops as the evaluation mesh becomes more dissimilar from the source, but \ours remains high generalization performance across all meshes.

\subsubsection{Zero-shot Cross-Category Results}

\begin{figure}[ht]
\begin{center}
\includegraphics[width=8cm]{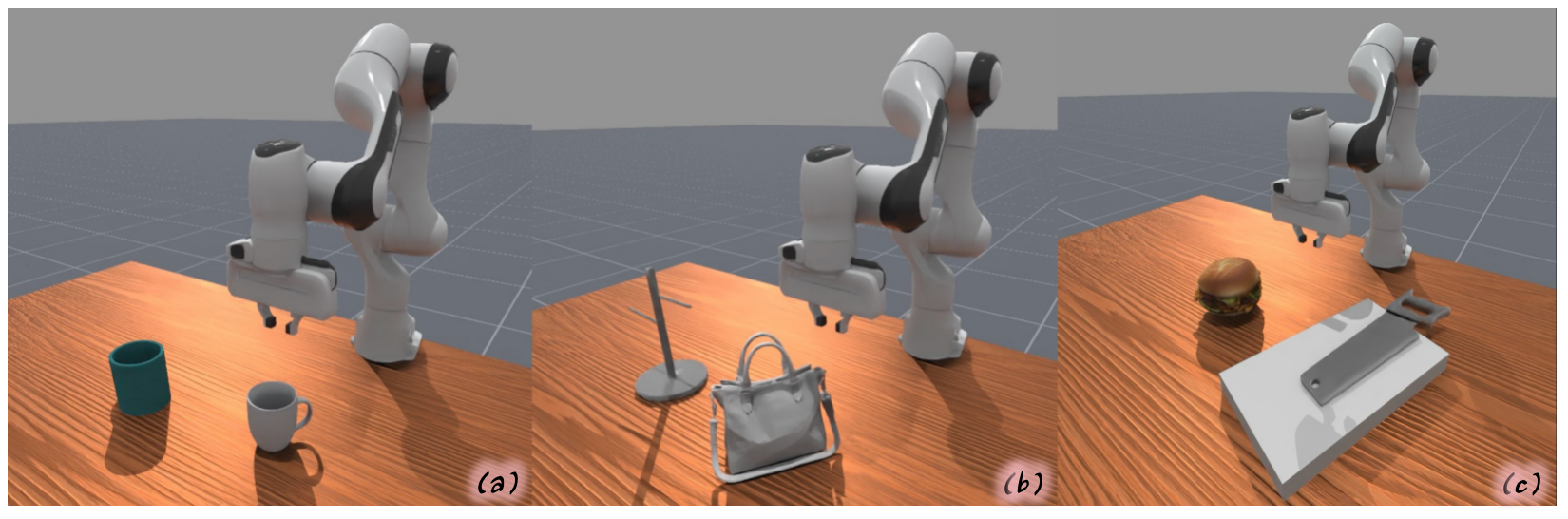}
\end{center}
\caption{Simulative cross-category tasks: (a) Mug Pouring, (b) Handbag Hanging, (c) Saw Cutting.}
\label{fig:sim_cross}
\end{figure}

By establishing correspondences between keypoints, AffordGen can repurpose source demonstrations to generate training data for cross-category objects. This synthetically generated data can then be used directly to train policies on novel object categories, as long as they share the same functional affordance. To showcase such capability, we conducted 3 zero-shot cross-category policy learning experiments, namely, \textit{mug pouring} (from teapot pouring), \textit{handbag hanging} (from mug hanging), and \textit{saw cutting} (from knife cutting). The simulation cross-category tasks are shown in Figure~\ref{fig:sim_cross}.



As shown in Table~\ref{tab:cross_cat}, \ours demonstrates impressive results in generating effective training data even for out-of-category objects. It is the only method to achieve a meaningful non-zero success rate on these new objects.
\begin{table*}[htbp]
\caption{Comparison of different methods across real-world tasks.}
\centering
\renewcommand{\arraystretch}{1.3}
\setlength{\tabcolsep}{6pt} 

\resizebox{0.7\textwidth}{!}{
\begin{tabular}{c cc cc cc cc}
\toprule
\multirow{2}{*}{\#Mesh $\times$ \#Demo} 
& \multicolumn{2}{c}{\textbf{Teapot Pouring}} 
& \multicolumn{2}{c}{\textbf{Mug Hanging}} 
& \multicolumn{2}{c}{\textbf{Knife Cutting}} 
& \multicolumn{2}{c}{\textbf{Shoe Organizing}} \\
\cmidrule(lr){2-3}
\cmidrule(lr){4-5}
\cmidrule(lr){6-7}
\cmidrule(lr){8-9}
& Source & Unseen
& Source & Unseen
& Source & Unseen
& Source & Unseen \\
\midrule
DemoGen (1$\times$1000) & $ \textbf{14/27} $ & $ 2/162 $ & $\textbf{20/27}$ & $74/162$ &  $ \textbf{25/27} $  & $47 / 108 $ & $13/20$ &  $24/60$ \\
CPGen (1000$\times$1) & $ 10/27 $ & $ 15/162 $ & $19/27$ & $69/162$ & $ 23/27 $ & $ 88/108 $ & $\textbf{18/20}$ & $30/60$ \\
AffordGen (100$\times$10) & $13/27$ & $ \textbf{74/162} $ & $19/27$ & $\textbf{107/162}$ & $23/27$ & $\textbf{96/108}$ & $11/20$&  $\textbf{45/60}$ \\
FUNCTO & $10/27$ & $ 50/162 $ & $ 7/27 $ & $48 /162 $ & $21/27$ & $61/108$ & $15/20$& $19/60$ \\
\bottomrule
\end{tabular}
} 

\label{tab:real_incat}
\end{table*}

\begin{table*}[htb]
  \centering
  
    \resizebox{0.9\linewidth}{!}{%
    \begin{tabular}{lcccccc}
    
    \toprule
    & \multicolumn{3}{c}{Simulation} & \multicolumn{3}{c}{Real-World} \\
    
    \cmidrule(lr){2-4} \cmidrule(lr){5-7}
    
    Algorithm / Tasks & 
    Teapot-Mug & Mug-Handbag & Knife-Saw &
    Teapot-Mug & Mug-Handbag & Knife-Saw \\
    
    \midrule
    
    \ours (ours) & 
    \textbf{55.00\%}  $\pm$ 9.10\% & \textbf{83.07\%}  $\pm$ 1.32\% & \textbf{40.22\%}  $\pm$ 7.28\% &
    \textbf{14 / 27} & \textbf{7 / 12} & \textbf{9 / 27} \\
    
    CPGen & 
    2.70\% $\pm$ 2.50\%  & 0.67\% $\pm$ 0.50\%  & 1.11\% $\pm$ 1.00\% &
    3 / 27 & 0 / 12 & 1 / 27 \\
    
    DemoGen & 
    0.70\% $\pm$ 0.90\%  & 0.27\% $\pm$ 0.38\%  & 1.56\% $\pm$ 0.38\% &
    0 / 27 & 0 / 12 & 1 / 27 \\
    
    \bottomrule
  \end{tabular}
  }
  \caption{Comparison of success rates on cross-category generalization tasks under simulation and real-world settings.}
  \label{tab:cross_cat}
\end{table*}

\subsection{Real World Experiments}

\subsubsection{Experiment Settings}

Real-world data generation is the most meaningful use case of \ours. While it is tedious and costly to collect data on thousands of different spatial relationships, it would be prohibitive to do so on thousands of different real objects.

To demonstrate the effectiveness of \ours in real-world generalization, we conduct the four simulation tasks in the real world. For each task, \textbf{10} expert demonstrations are collected to generate \textbf{1000} training demonstrations. Following DemoGen, we increase the number of real demonstrations to avoid overfitting. After \textbf{training solely on the generated data}, we evaluate the learned policies on a variety of previously unseen real objects. 
\begin{figure}[htbp]
    \centering
    \begin{subfigure}{0.21\textwidth}
        \centering
        \includegraphics[width=\linewidth]{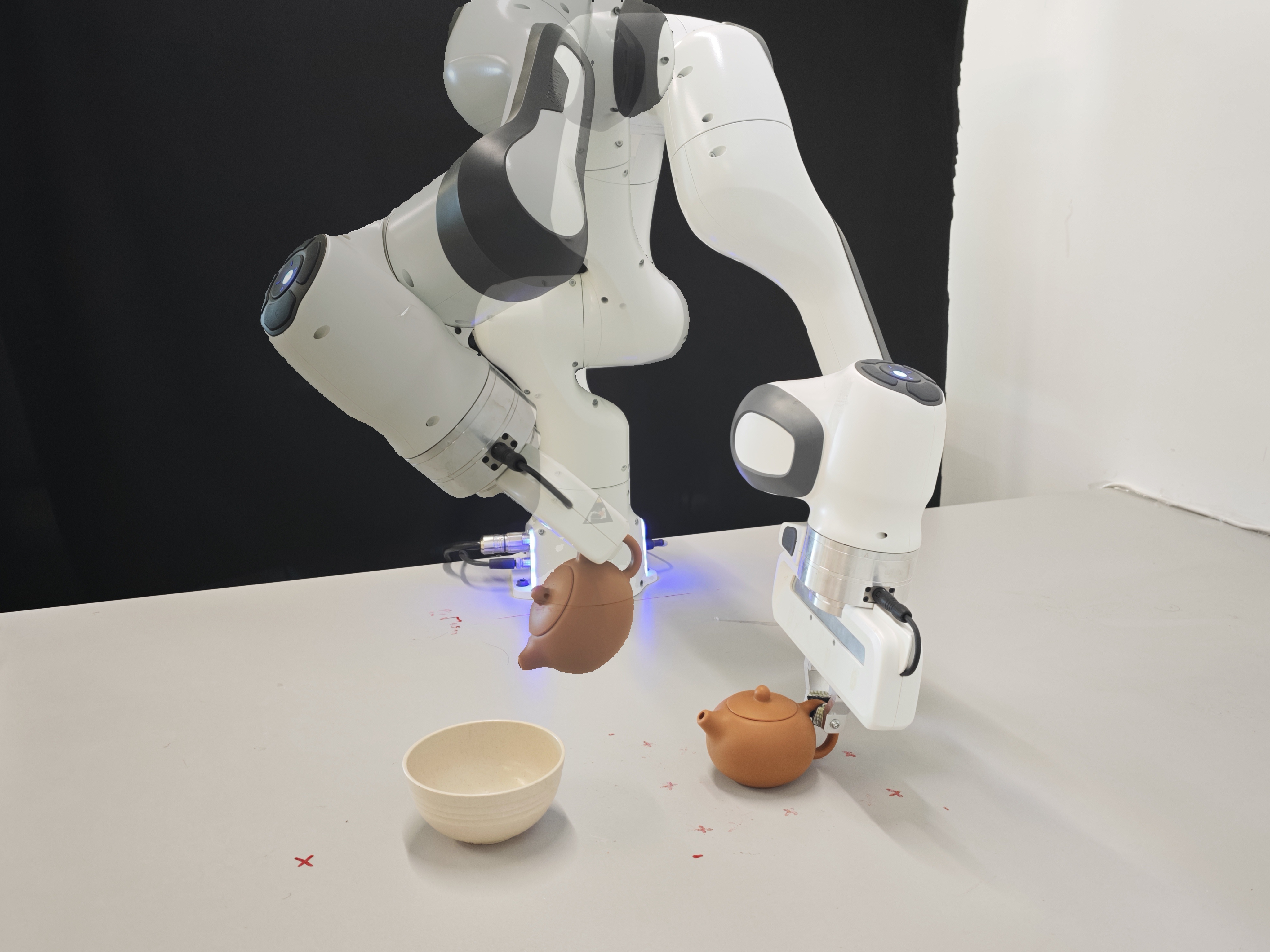}
        \caption{Teapot Pouring}
        \label{fig:real_teapot}
    \end{subfigure}
    \begin{subfigure}{0.21\textwidth}
        \centering
        \includegraphics[width=\linewidth]{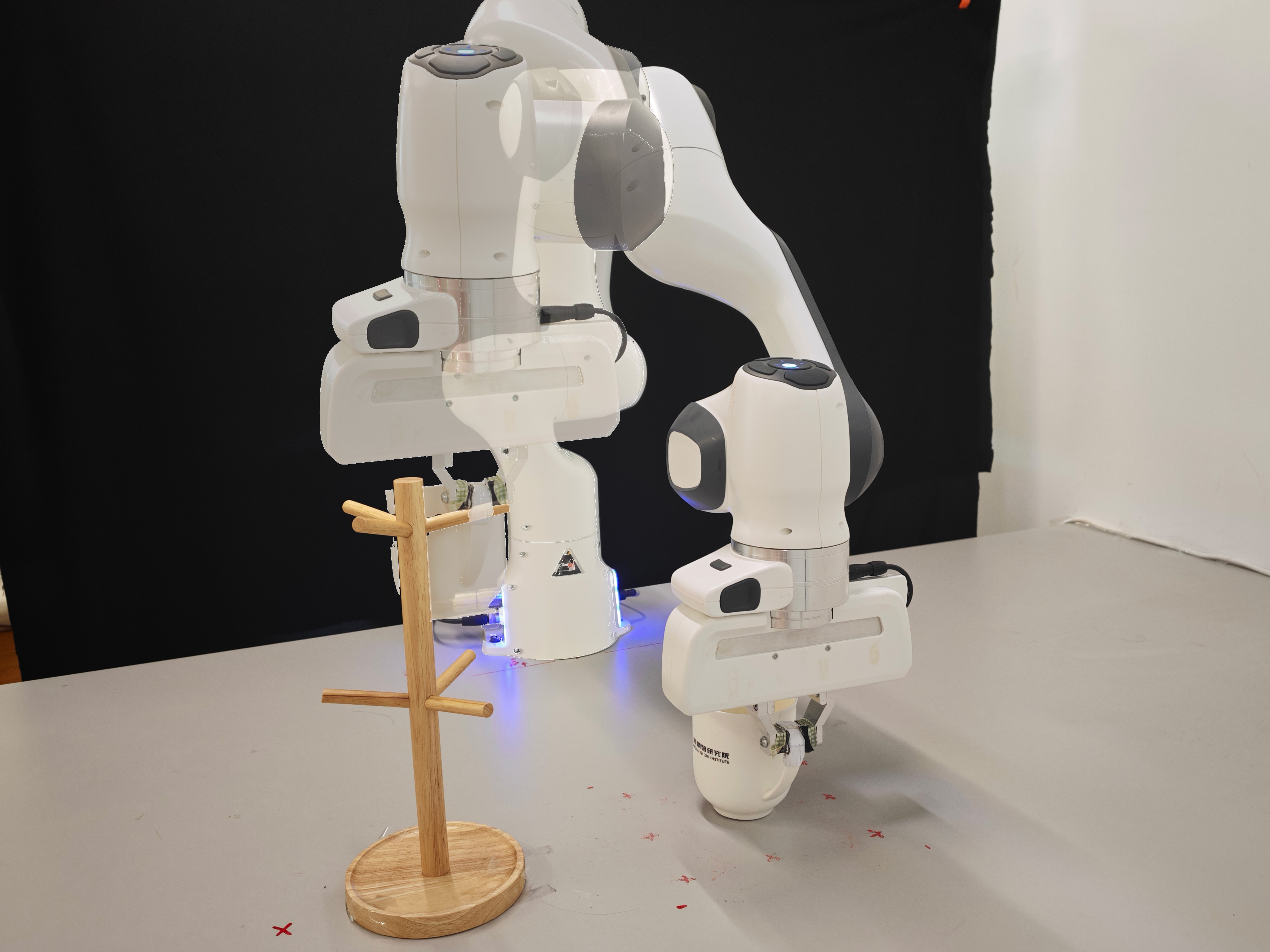}
        \caption{Mug Hanging}
        \label{fig:real_mug}
    \end{subfigure}
    \vspace{0.5cm} 
    
    \begin{subfigure}{0.21\textwidth}
        \centering
        \includegraphics[width=\linewidth]{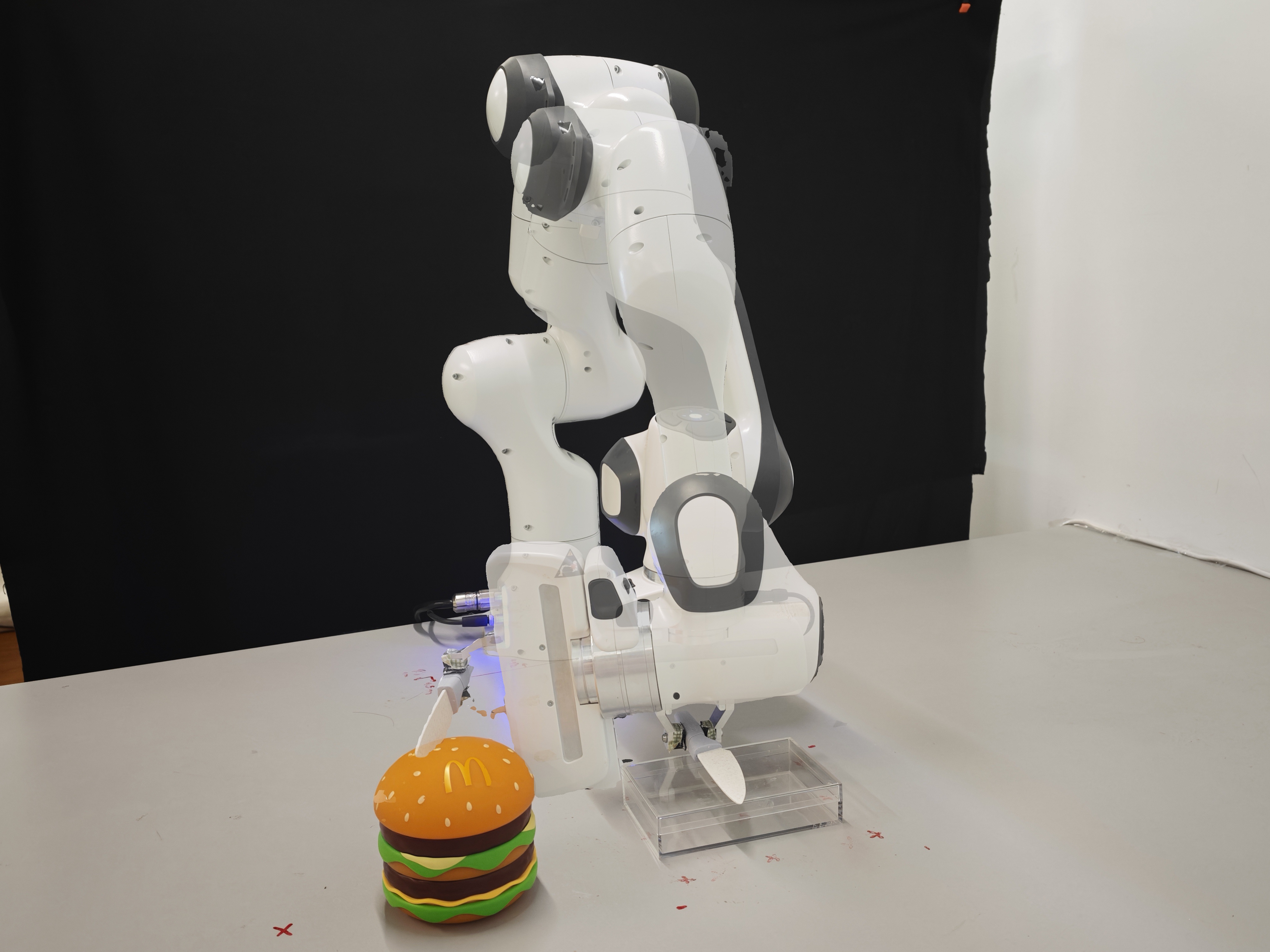}
        \caption{Knife Cutting}
        \label{fig:real_knife}
    \end{subfigure}
    \begin{subfigure}{0.21\textwidth}
        \centering
        \includegraphics[width=\linewidth]{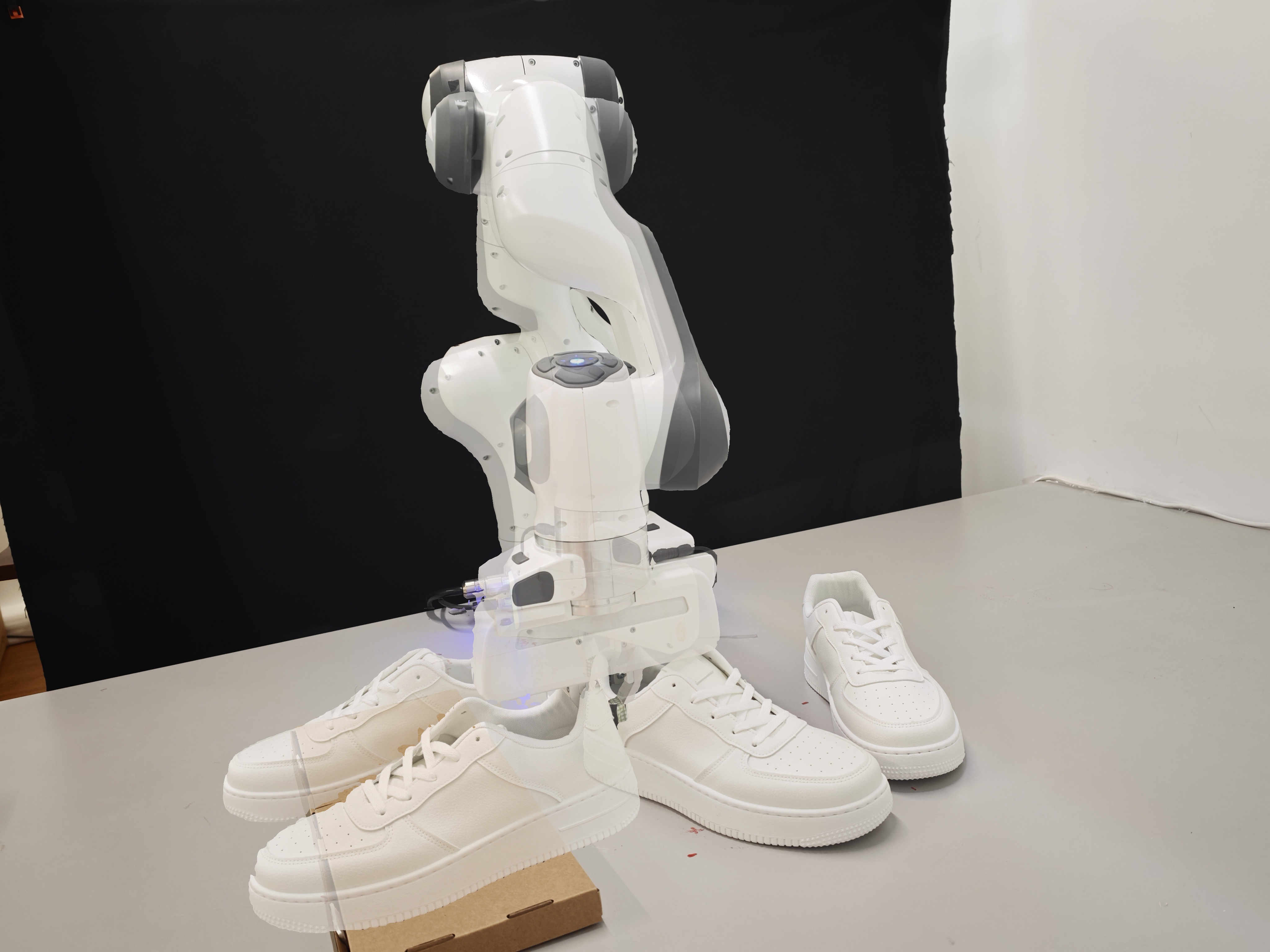}
        \caption{Shoe Organizing}
        \label{fig:real_shoe}
    \end{subfigure}
    
    \caption{Real-World experiments setup}
    \label{fig:real_tasks}
\end{figure}

In real-world experiments, we include another planning-based method, FUNCTO \cite{Tang2025FUNCTOFO}. FUNCTO serves as a representative algorithm based on keypoint correspondence. Similar to AffordGen, FUNCTO generates manipulation trajectories for new objects through semantic mapping. Leveraging large language models and visual foundation models, it performs path planning according to keypoint correspondences. We include FUNCTO to highlight the differences between AffordGen and affordance-based open-loop planning algorithms, and demonstrate how AffordGen overcomes the limitations of such approaches.
\subsubsection{In-category Generalization Results}

We first conduct in-category experiments on the source object and a sufficiently diverse set of unseen test objects. The task setups are illustrated in Figure \ref{fig:real_tasks}, while the detailed evaluation setting can be found in Appendix~\ref{app:exp_details}. 

For a fair comparison, we evaluated all test objects under a fixed set of poses: 27 poses (9 positions x 3 orientations) for most tasks, and 10 poses (2 positions x 5 orientation) specifically for the shoe task due to its long-horizon nature.

The results are presented in Table \ref{tab:real_incat}. We see a similar pattern to the simulation tasks: with only a small amount of real-world expert data, DemoGen, CPGen, and AffordGen all achieve high success rates across the entire workspace on the source mesh, while CPGen significantly outperforms DemoGen in most unseen tests, and \ours further beats both baselines.

FUNCTO's performance heavily depends on the selection of correspondence points, which is often compromised by factors such as occlusion of key parts and large view perspective differences between source and target objects. When the key regions of the target object remain clearly visible, such as knife cutting, FUNCTO maintains a relatively high success rate. However, in scenarios with large orientation variations and occlusion, such as mug hanging, FUNCTO yields the worst performance among all baseline methods. Benefiting from the 3D space correspondences and training on large-scale generated data, AffordGen implicitly learns the relationships between affording points, function points, and object shapes, effectively resolving the keypoint occlusion issues common in planner-based methods.

\subsubsection{Zero-shot Cross-category Results}
\begin{figure}[ht]
\begin{center}
\includegraphics[width=8cm]{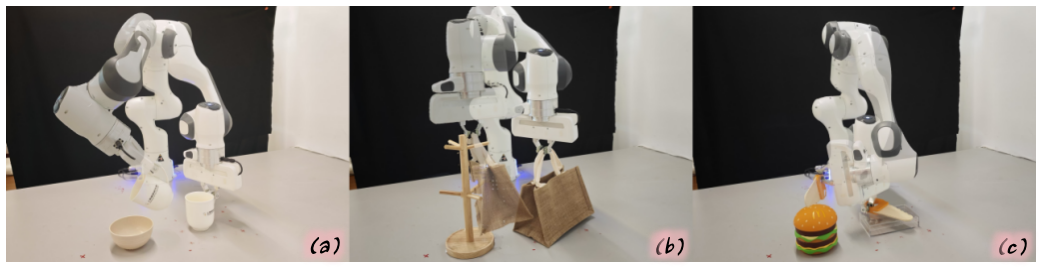}
\end{center}
\caption{Real cross-category tasks settings: (a) Mug Pouring, (b) Handbag Hanging, (c) Saw Cutting.}
\label{fig:real_cross}
\vspace{-0.5cm}
\end{figure}

We establish a real-world setup identical to the simulation for cross-category testing, as shown in Figure~\ref{fig:real_cross}. The experimental results in Table \ref{tab:cross_cat} demonstrate that \ours can effectively generate cross-category object manipulation data in the real world, thereby further expanding real-world robot capabilities with low cost.



\section{Conclusion}
\ours addresses the data scarcity and generalization problem in robotic learning. This work presents a new paradigm that leverages affordance correspondence as a generative source to scale minimal demonstrations into thousands of diverse, semantically-grounded, and full 6D trajectories across object categories. Policies trained on this synthetic dataset achieve robust, closed-loop control and demonstrate strong generalization to truly unseen objects, indicating great potential for large-scale real-world applications.
\clearpage
\newpage
\section*{Acknowlegement}
This work is supported by Tsinghua University-Keystone Electrical (Zhejiang) Co., id Joint Research Center for Embodied Multimodal Artificial Intelligence(JCEMAl).
{
    \small
    \bibliographystyle{ieeenat_fullname}
    \bibliography{main}

@article{Chi2023DiffusionPV,
  title={Diffusion Policy: Visuomotor Policy Learning via Action Diffusion},
  author={Cheng Chi and Siyuan Feng and Yilun Du and Zhenjia Xu and Eric Cousineau and Benjamin Burchfiel and Shuran Song},
  journal={ArXiv},
  year={2023},
  volume={abs/2303.04137},
  url={https://api.semanticscholar.org/CorpusID:257378658}
}

@inproceedings{Wang2023MimicPlayLI,
  title={MimicPlay: Long-Horizon Imitation Learning by Watching Human Play},
  author={Chen Wang and Linxi (Jim) Fan and Jiankai Sun and Ruohan Zhang and Li Fei-Fei and Danfei Xu and Yuke Zhu and Anima Anandkumar},
  booktitle={Conference on Robot Learning},
  year={2023},
  url={https://api.semanticscholar.org/CorpusID:257205825}
}

@article{Liu2024RDT1BAD,
  title={RDT-1B: a Diffusion Foundation Model for Bimanual Manipulation},
  author={Songming Liu and Lingxuan Wu and Bangguo Li and Hengkai Tan and Huayu Chen and Zhengyi Wang and Ke Xu and Hang Su and Jun Zhu},
  journal={ArXiv},
  year={2024},
  volume={abs/2410.07864},
  url={https://api.semanticscholar.org/CorpusID:273233386}
}

@article{Lin2024DataSL,
  title={Data Scaling Laws in Imitation Learning for Robotic Manipulation},
  author={Fanqi Lin and Yingdong Hu and Pingyue Sheng and Chuan Wen and Jiacheng You and Yang Gao},
  journal={ArXiv},
  year={2024},
  volume={abs/2410.18647},
  url={https://api.semanticscholar.org/CorpusID:273549927}
}

@article{Xue2025DemoGenSD,
  title={DemoGen: Synthetic Demonstration Generation for Data-Efficient Visuomotor Policy Learning},
  author={Zhengrong Xue and Shuying Deng and Zhenyang Chen and Yixuan Wang and Zhecheng Yuan and Huazhe Xu},
  journal={ArXiv},
  year={2025},
  volume={abs/2502.16932},
  url={https://api.semanticscholar.org/CorpusID:276575314}
}

@article{Yuan2023RLViGenAR,
  title={RL-ViGen: A Reinforcement Learning Benchmark for Visual Generalization},
  author={Zhecheng Yuan and Sizhe Yang and Pu Hua and Cancer Suk Chul Chang and Kaizhe Hu and Xiaolong Wang and Huazhe Xu},
  journal={ArXiv},
  year={2023},
  volume={abs/2307.10224},
  url={https://api.semanticscholar.org/CorpusID:259991209}
}

@inproceedings{Ju2024RoboABCAG,
  title={Robo-ABC: Affordance Generalization Beyond Categories via Semantic Correspondence for Robot Manipulation},
  author={Yuanchen Ju and Kaizhe Hu and Guowei Zhang and Gu Zhang and Mingrun Jiang and Huazhe Xu},
  booktitle={European Conference on Computer Vision},
  year={2024},
  url={https://api.semanticscholar.org/CorpusID:266999371}
}

@article{zhu2024densematcher,
  title={Densematcher: Learning 3d semantic correspondence for category-level manipulation from a single demo},
  author={Zhu, Junzhe and Ju, Yuanchen and Zhang, Junyi and Wang, Muhan and Yuan, Zhecheng and Hu, Kaizhe and Xu, Huazhe},
  journal={arXiv preprint arXiv:2412.05268},
  year={2024}
}

@article{Tang2025FUNCTOFO,
  title={FUNCTO: Function-Centric One-Shot Imitation Learning for Tool Manipulation},
  author={Chao Tang and Anxing Xiao and Yuhong Deng and Tianrun Hu and Wenlong Dong and Hanbo Zhang and David Hsu and Hong Zhang},
  journal={ArXiv},
  year={2025},
  volume={abs/2502.11744},
  url={https://api.semanticscholar.org/CorpusID:276408407}
}

@inproceedings{zhang2024omni6dpose,
  title={Omni6dpose: A benchmark and model for universal 6d object pose estimation and tracking},
  author={Zhang, Jiyao and Huang, Weiyao and Peng, Bo and Wu, Mingdong and Hu, Fei and Chen, Zijian and Zhao, Bo and Dong, Hao},
  booktitle={European Conference on Computer Vision},
  pages={199--216},
  year={2024},
  organization={Springer}
}

@misc{hunyuan3d22025tencent,
    title={Hunyuan3D 2.0: Scaling Diffusion Models for High Resolution Textured 3D Assets Generation},
    author={Tencent Hunyuan3D Team},
    year={2025},
    eprint={2501.12202},
    archivePrefix={arXiv},
    primaryClass={cs.CV}
}

@article{taomaniskill3,
  title={ManiSkill3: GPU Parallelized Robotics Simulation and Rendering for Generalizable Embodied AI},
  author={Stone Tao and Fanbo Xiang and Arth Shukla and Yuzhe Qin and Xander Hinrichsen and Xiaodi Yuan and Chen Bao and Xinsong Lin and Yulin Liu and Tse-kai Chan and Yuan Gao and Xuanlin Li and Tongzhou Mu and Nan Xiao and Arnav Gurha and Viswesh Nagaswamy Rajesh and Yong Woo Choi and Yen-Ru Chen and Zhiao Huang and Roberto Calandra and Rui Chen and Shan Luo and Hao Su},
  journal = {Robotics: Science and Systems},
  year={2025},
}

@article{ravi2024sam2,
  title={SAM 2: Segment Anything in Images and Videos},
  author={Ravi, Nikhila and Gabeur, Valentin and Hu, Yuan-Ting and Hu, Ronghang and Ryali, Chaitanya and Ma, Tengyu and Khedr, Haitham and R{\"a}dle, Roman and Rolland, Chloe and Gustafson, Laura and Mintun, Eric and Pan, Junting and Alwala, Kalyan Vasudev and Carion, Nicolas and Wu, Chao-Yuan and Girshick, Ross and Doll{\'a}r, Piotr and Feichtenhofer, Christoph},
  journal={arXiv preprint arXiv:2408.00714},
  url={https://arxiv.org/abs/2408.00714},
  year={2024}
}

@misc{ze20243d,
      title={3D Diffusion Policy: Generalizable Visuomotor Policy Learning via Simple 3D Representations}, 
      author={Yanjie Ze and Gu Zhang and Kangning Zhang and Chenyuan Hu and Muhan Wang and Huazhe Xu},
      year={2024},
      eprint={2403.03954},
      archivePrefix={arXiv},
      primaryClass={cs.RO},
      url={https://arxiv.org/abs/2403.03954}, 
}

@misc{mandlekar2023mimicgen,
      title={MimicGen: A Data Generation System for Scalable Robot Learning using Human Demonstrations}, 
      author={Ajay Mandlekar and Soroush Nasiriany and Bowen Wen and Iretiayo Akinola and Yashraj Narang and Linxi Fan and Yuke Zhu and Dieter Fox},
      year={2023},
      eprint={2310.17596},
      archivePrefix={arXiv},
      primaryClass={cs.RO},
      url={https://arxiv.org/abs/2310.17596}, 
}

@misc{garrett2024skillmimicgen,
      title={SkillMimicGen: Automated Demonstration Generation for Efficient Skill Learning and Deployment}, 
      author={Caelan Garrett and Ajay Mandlekar and Bowen Wen and Dieter Fox},
      year={2024},
      eprint={2410.18907},
      archivePrefix={arXiv},
      primaryClass={cs.RO},
      url={https://arxiv.org/abs/2410.18907}, 
}

@misc{jiang2025dexmimicgen,
      title={DexMimicGen: Automated Data Generation for Bimanual Dexterous Manipulation via Imitation Learning}, 
      author={Zhenyu Jiang and Yuqi Xie and Kevin Lin and Zhenjia Xu and Weikang Wan and Ajay Mandlekar and Linxi Fan and Yuke Zhu},
      year={2025},
      eprint={2410.24185},
      archivePrefix={arXiv},
      primaryClass={cs.RO},
      url={https://arxiv.org/abs/2410.24185}, 
}

@misc{lin2025constraint,
      title={Constraint-Preserving Data Generation for Visuomotor Policy Learning}, 
      author={Kevin Lin and Varun Ragunath and Andrew McAlinden and Aaditya Prasad and Jimmy Wu and Yuke Zhu and Jeannette Bohg},
      year={2025},
      eprint={2508.03944},
      archivePrefix={arXiv},
      primaryClass={cs.RO},
      url={https://arxiv.org/abs/2508.03944}, 
}

@misc{wang2024gensim,
      title={GenSim: Generating Robotic Simulation Tasks via Large Language Models}, 
      author={Lirui Wang and Yiyang Ling and Zhecheng Yuan and Mohit Shridhar and Chen Bao and Yuzhe Qin and Bailin Wang and Huazhe Xu and Xiaolong Wang},
      year={2024},
      eprint={2310.01361},
      archivePrefix={arXiv},
      primaryClass={cs.LG},
      url={https://arxiv.org/abs/2310.01361}, 
}

@misc{hua2024gensim2,
      title={GenSim2: Scaling Robot Data Generation with Multi-modal and Reasoning LLMs}, 
      author={Pu Hua and Minghuan Liu and Annabella Macaluso and Yunfeng Lin and Weinan Zhang and Huazhe Xu and Lirui Wang},
      year={2024},
      eprint={2410.03645},
      archivePrefix={arXiv},
      primaryClass={cs.RO},
      url={https://arxiv.org/abs/2410.03645}, 
}

@misc{wang2024robogen,
      title={RoboGen: Towards Unleashing Infinite Data for Automated Robot Learning via Generative Simulation}, 
      author={Yufei Wang and Zhou Xian and Feng Chen and Tsun-Hsuan Wang and Yian Wang and Katerina Fragkiadaki and Zackory Erickson and David Held and Chuang Gan},
      year={2024},
      eprint={2311.01455},
      archivePrefix={arXiv},
      primaryClass={cs.RO},
      url={https://arxiv.org/abs/2311.01455}, 
}

@misc{chen2023genaug,
      title={GenAug: Retargeting behaviors to unseen situations via Generative Augmentation}, 
      author={Zoey Chen and Sho Kiami and Abhishek Gupta and Vikash Kumar},
      year={2023},
      eprint={2302.06671},
      archivePrefix={arXiv},
      primaryClass={cs.RO},
      url={https://arxiv.org/abs/2302.06671}, 
}

@misc{nasiriany2024robocasa,
      title={RoboCasa: Large-Scale Simulation of Everyday Tasks for Generalist Robots}, 
      author={Soroush Nasiriany and Abhiram Maddukuri and Lance Zhang and Adeet Parikh and Aaron Lo and Abhishek Joshi and Ajay Mandlekar and Yuke Zhu},
      year={2024},
      eprint={2406.02523},
      archivePrefix={arXiv},
      primaryClass={cs.RO},
      url={https://arxiv.org/abs/2406.02523}, 
}

@misc{yu2023scaling,
      title={Scaling Robot Learning with Semantically Imagined Experience}, 
      author={Tianhe Yu and Ted Xiao and Austin Stone and Jonathan Tompson and Anthony Brohan and Su Wang and Jaspiar Singh and Clayton Tan and Dee M and Jodilyn Peralta and Brian Ichter and Karol Hausman and Fei Xia},
      year={2023},
      eprint={2302.11550},
      archivePrefix={arXiv},
      primaryClass={cs.RO},
      url={https://arxiv.org/abs/2302.11550}, 
}

@misc{mandi2023cacti,
      title={CACTI: A Framework for Scalable Multi-Task Multi-Scene Visual Imitation Learning}, 
      author={Zhao Mandi and Homanga Bharadhwaj and Vincent Moens and Shuran Song and Aravind Rajeswaran and Vikash Kumar},
      year={2023},
      eprint={2212.05711},
      archivePrefix={arXiv},
      primaryClass={cs.RO},
      url={https://arxiv.org/abs/2212.05711}, 
}

@inproceedings{pan2025omnimanip,
  title={Omnimanip: Towards general robotic manipulation via object-centric interaction primitives as spatial constraints},
  author={Pan, Mingjie and Zhang, Jiyao and Wu, Tianshu and Zhao, Yinghao and Gao, Wenlong and Dong, Hao},
  booktitle={Proceedings of the Computer Vision and Pattern Recognition Conference},
  pages={17359--17369},
  year={2025}
}

@misc{oquab2024dinov2,
      title={DINOv2: Learning Robust Visual Features without Supervision}, 
      author={Maxime Oquab and Timothée Darcet and Théo Moutakanni and Huy Vo and Marc Szafraniec and Vasil Khalidov and Pierre Fernandez and Daniel Haziza and Francisco Massa and Alaaeldin El-Nouby and Mahmoud Assran and Nicolas Ballas and Wojciech Galuba and Russell Howes and Po-Yao Huang and Shang-Wen Li and Ishan Misra and Michael Rabbat and Vasu Sharma and Gabriel Synnaeve and Hu Xu and Hervé Jegou and Julien Mairal and Patrick Labatut and Armand Joulin and Piotr Bojanowski},
      year={2024},
      eprint={2304.07193},
      archivePrefix={arXiv},
      primaryClass={cs.CV},
      url={https://arxiv.org/abs/2304.07193}, 
}

@inproceedings{shoemake1985slerp,
author = {Shoemake, Ken},
title = {Animating rotation with quaternion curves},
year = {1985},
isbn = {0897911660},
publisher = {Association for Computing Machinery},
address = {New York, NY, USA},
url = {https://doi.org/10.1145/325334.325242},
doi = {10.1145/325334.325242},
booktitle = {Proceedings of the 12th Annual Conference on Computer Graphics and Interactive Techniques},
pages = {245–254},
numpages = {10},
series = {SIGGRAPH '85}
}

@misc{trilbmteam2025careful,
      title={A Careful Examination of Large Behavior Models for Multitask Dexterous Manipulation}, 
      author={TRI LBM Team and Jose Barreiros and Andrew Beaulieu and Aditya Bhat and Rick Cory and Eric Cousineau and Hongkai Dai and Ching-Hsin Fang and Kunimatsu Hashimoto and Muhammad Zubair Irshad and Masha Itkina and Naveen Kuppuswamy and Kuan-Hui Lee and Katherine Liu and Dale McConachie and Ian McMahon and Haruki Nishimura and Calder Phillips-Grafflin and Charles Richter and Paarth Shah and Krishnan Srinivasan and Blake Wulfe and Chen Xu and Mengchao Zhang and Alex Alspach and Maya Angeles and Kushal Arora and Vitor Campagnolo Guizilini and Alejandro Castro and Dian Chen and Ting-Sheng Chu and Sam Creasey and Sean Curtis and Richard Denitto and Emma Dixon and Eric Dusel and Matthew Ferreira and Aimee Goncalves and Grant Gould and Damrong Guoy and Swati Gupta and Xuchen Han and Kyle Hatch and Brendan Hathaway and Allison Henry and Hillel Hochsztein and Phoebe Horgan and Shun Iwase and Donovon Jackson and Siddharth Karamcheti and Sedrick Keh and Joseph Masterjohn and Jean Mercat and Patrick Miller and Paul Mitiguy and Tony Nguyen and Jeremy Nimmer and Yuki Noguchi and Reko Ong and Aykut Onol and Owen Pfannenstiehl and Richard Poyner and Leticia Priebe Mendes Rocha and Gordon Richardson and Christopher Rodriguez and Derick Seale and Michael Sherman and Mariah Smith-Jones and David Tago and Pavel Tokmakov and Matthew Tran and Basile Van Hoorick and Igor Vasiljevic and Sergey Zakharov and Mark Zolotas and Rares Ambrus and Kerri Fetzer-Borelli and Benjamin Burchfiel and Hadas Kress-Gazit and Siyuan Feng and Stacie Ford and Russ Tedrake},
      year={2025},
      eprint={2507.05331},
      archivePrefix={arXiv},
      primaryClass={cs.RO},
      url={https://arxiv.org/abs/2507.05331}, 
}

@article{florence2018dense,
  title={Dense object nets: Learning dense visual object descriptors by and for robotic manipulation},
  author={Florence, Peter R and Manuelli, Lucas and Tedrake, Russ},
  journal={arXiv preprint arXiv:1806.08756},
  year={2018}
}

@inproceedings{manuelli2019kpam,
  title={kpam: Keypoint affordances for category-level robotic manipulation},
  author={Manuelli, Lucas and Gao, Wei and Florence, Peter and Tedrake, Russ},
  booktitle={The International Symposium of Robotics Research},
  pages={132--157},
  year={2019},
  organization={Springer}
}

@inproceedings{simeonov2022neural,
  title={Neural descriptor fields: Se (3)-equivariant object representations for manipulation},
  author={Simeonov, Anthony and Du, Yilun and Tagliasacchi, Andrea and Tenenbaum, Joshua B and Rodriguez, Alberto and Agrawal, Pulkit and Sitzmann, Vincent},
  booktitle={2022 International Conference on Robotics and Automation (ICRA)},
  pages={6394--6400},
  year={2022},
  organization={IEEE}
}

@article{tang2023emergent,
  title={Emergent correspondence from image diffusion},
  author={Tang, Luming and Jia, Menglin and Wang, Qianqian and Phoo, Cheng Perng and Hariharan, Bharath},
  journal={Advances in Neural Information Processing Systems},
  volume={36},
  pages={1363--1389},
  year={2023}
}

@article{zhang2023tale,
  title={A tale of two features: Stable diffusion complements dino for zero-shot semantic correspondence},
  author={Zhang, Junyi and Herrmann, Charles and Hur, Junhwa and Polania Cabrera, Luisa and Jampani, Varun and Sun, Deqing and Yang, Ming-Hsuan},
  journal={Advances in Neural Information Processing Systems},
  volume={36},
  pages={45533--45547},
  year={2023}
}

@inproceedings{wu2025afforddp,
  title={Afforddp: Generalizable diffusion policy with transferable affordance},
  author={Wu, Shijie and Zhu, Yihang and Huang, Yunao and Zhu, Kaizhen and Gu, Jiayuan and Yu, Jingyi and Shi, Ye and Wang, Jingya},
  booktitle={Proceedings of the Computer Vision and Pattern Recognition Conference},
  pages={6971--6980},
  year={2025}
}

@article{kuang2024ram,
  title={Ram: Retrieval-based affordance transfer for generalizable zero-shot robotic manipulation},
  author={Kuang, Yuxuan and Ye, Junjie and Geng, Haoran and Mao, Jiageng and Deng, Congyue and Guibas, Leonidas and Wang, He and Wang, Yue},
  journal={arXiv preprint arXiv:2407.04689},
  year={2024}
}

@inproceedings{srirama2024hrp,
  title     = {HRP: Human Affordances for Robotic Pre-Training},
  author    = {Srirama, Mohan Kumar and Dasari, Sudeep and Bahl, Shikhar and Gupta, Abhinav},
  booktitle = {Robotics: Science and Systems (RSS)},
  year      = {2024},
  address   = {Delft, Netherlands}
}

@article{tang2025mimicfunc,
  title={Mimicfunc: Imitating tool manipulation from a single human video via functional correspondence},
  author={Tang, Chao and Xiao, Anxing and Deng, Yuhong and Hu, Tianrun and Dong, Wenlong and Zhang, Hanbo and Hsu, David and Zhang, Hong},
  journal={arXiv preprint arXiv:2508.13534},
  year={2025}
}

@article{hu2024generalizable,
  title={Generalizable visual imitation learning with stem-like convergent observation through diffusion inversion},
  author={Hu, Kaizhe and Rui, Zihang and He, Yao and Liu, Yuyao and Hua, Pu},
  journal={arXiv preprint arXiv:2411.04919},
  volume={1},
  year={2024}
}
}
\clearpage
\setcounter{page}{1}
\maketitlesupplementary

\section{Hyperparameters}
\label{app:hyper_param}
\subsection{Point Cloud Processing}
We follow the preprocessing pipeline for point cloud observations as outlined in DP3~\cite{ze20243d}. For simulation tasks, we directly apply Farthest Point Sampling (FPS) to downsample the point cloud to $1024$ points. For real-world tasks, we collect point clouds using a RealSense L515 camera at a depth image resolution of $1024 \times 768$ pixels. The collected point cloud is first clustered using DBSCAN to mitigate the influence of outliers, then downsampled to $1024$ points via FPS. In real-robot experiments, an $\epsilon$ of $1.0$ cm was used for the teapot, mug, and knife experiments, while an $\epsilon$ of $2.0$ cm was used for the shoe experiment.

\subsection{Policy Training and Evaluation}
For all simulation and real tasks, we train the policy for $1000$ epochs with a batch size of $512$. To stabilize the training process, we use the Adam optimizer and cosine learning rate scheduler with $500$ warmup steps. We set the observation horizon $T_o=2$, action prediction horizon $T_p=16$, and action execution horizon $T_a=6$ for the teapot, mug, and knife tasks. For long-horizon tasks like the shoe, we set $T_o=2$, $T_p=16$, and $T_a=16$ to avoid getting stuck.

\section{Experiment Details}
\label{app:exp_details}
\subsection{Task Description}
We summarize the four tasks (which are the same for both simulation and real-world setups) as follows:

\begin{itemize}
    \item \textbf{Teapot Pouring:} Grasp the handle, position the spout above the cup, and tilt beyond a threshold angle.
    \item \textbf{Mug Hanging:} Grasp the handle and hang the mug by threading its handle onto the rack.
    \item \textbf{Knife Cutting:} Grasp the handle and bring the blade into perpendicular contact with the object.
    \item \textbf{Shoe Aligning:} Grasp the heel and place both shoes on the rack in the same orientation.
\end{itemize}

For simulation tasks, we summarize the randomization range during evaluation as follows:
\begin{table}[!htbp]
\centering
\small 
\setlength{\tabcolsep}{4pt} 
\label{tab:task_parameters}
\begin{tabular}{ccccc}
\toprule
 & \textbf{Teapot} & \textbf{Mug} & \textbf{Knife} & \textbf{Shoe} \\
\midrule
\textbf{Position (cm)} & $20 \times 20$ & $10 \times 15$ & $20 \times 20$ & $15 \times 25$ \\
\textbf{Orientation (°)} & $[0, 180]$ & $[0, 180]$ & $[0, 90]$ & $[0, 120]$ \\
\bottomrule
\end{tabular}
\end{table}

\begin{figure}[ht]
\begin{center}
\includegraphics[width=6cm]{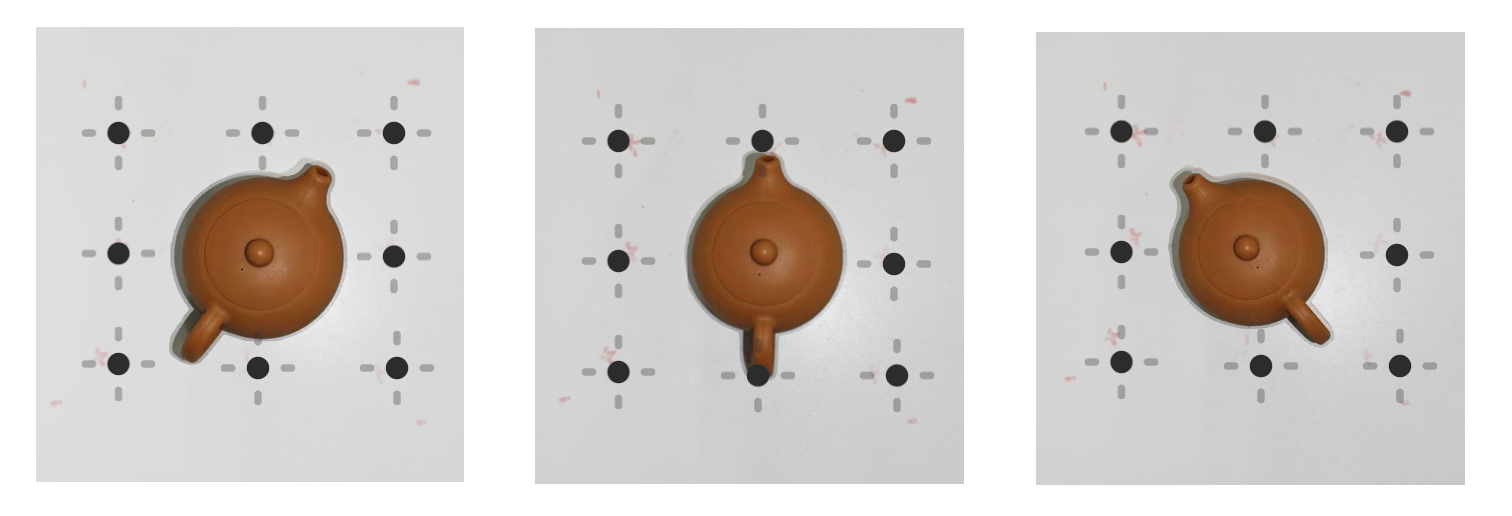}
\end{center}
\caption{$3 \times 3$ grid with three different orientations during real teapot, mug and knife evaluation.}
\label{fig:real_eval}
\end{figure}

For the real teapot, mug, and knife tasks, during testing, we placed each tested instance in a $3\times3$ grid, evaluating three orientations at each position in a total of $27$ rollouts for each eval instance as illustrated in Figure~\ref{fig:real_eval}.

\begin{figure}[ht]
\begin{center}
\includegraphics[width=8cm]{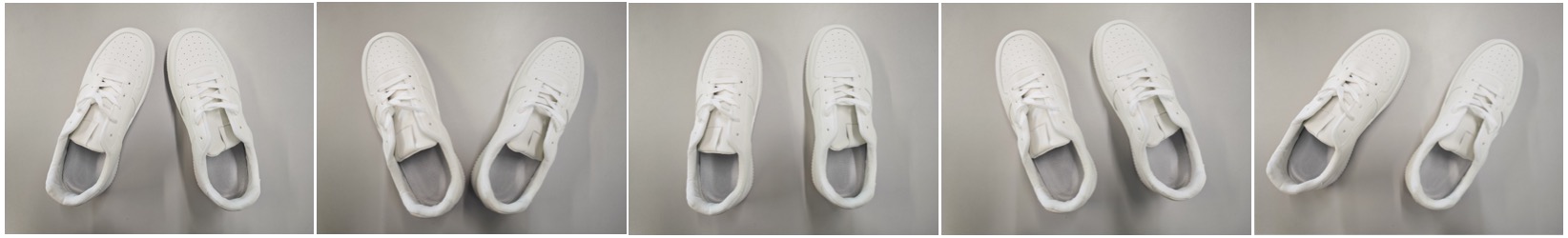}
\end{center}
\caption{Five pose configurations for real shoe evaluation.}
\label{fig:supp_shoe_config}
\end{figure}

For the real shoe task, we designed $5$ initial pose configurations, as shown in Figure~\ref{fig:supp_shoe_config}, for the
two shoes to be evaluated twice at varied positions, resulting in $10$ rollouts per instance. Note that the success times we report refer to the total number of successfully placed shoes, not the number of fully successful instances. Since each instance comprises both left and right shoes, the total number of times is $20$, which is twice the number of evaluation rollouts.

\subsection{Detailed Real-world Results}
We summarize the detailed success rates on each tested instance in the following sections. Note that each instance labeled 'a' serves as the source object for this task.
\subsubsection{Teapot Pouring}
\begin{figure}[H]
\begin{center}
\includegraphics[width=5cm]{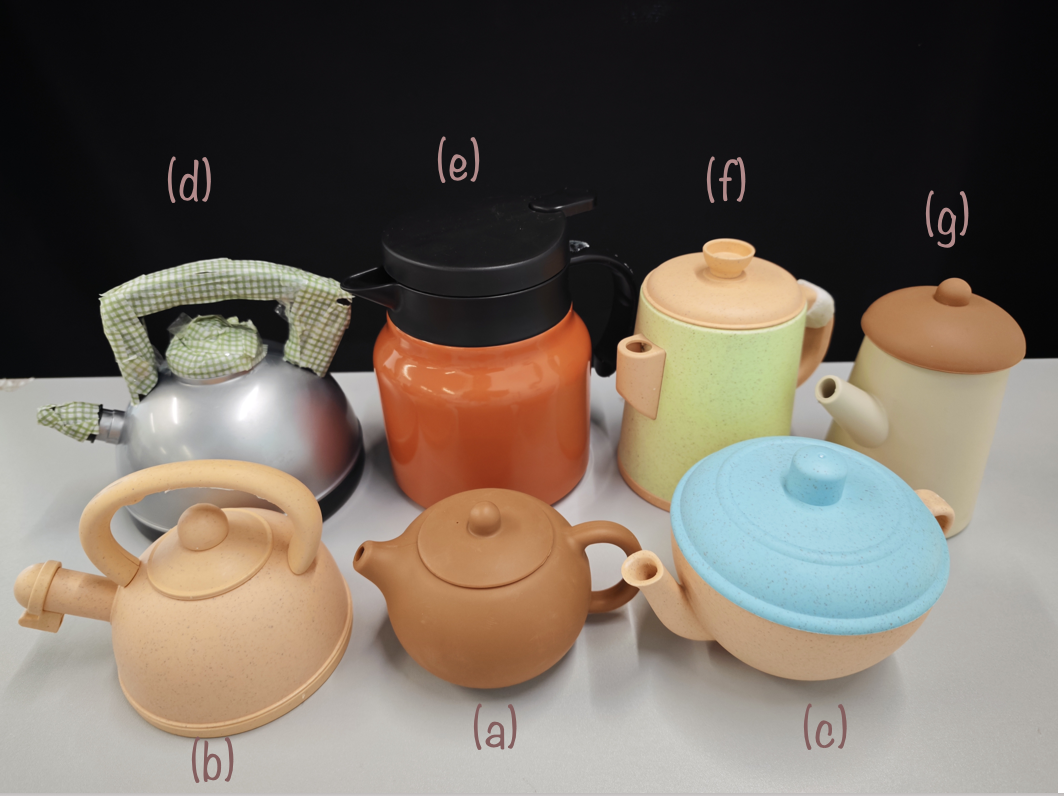}
\end{center}
\label{fig:supp_sr_teapot}
\caption{Teapot evaluation instances}
\end{figure}
\begin{table}[!ht]
\centering
\small 
\setlength{\tabcolsep}{3pt} 
\begin{tabular}{cccccccc}
\toprule
 & \textbf{a} & \textbf{b} & \textbf{c} & \textbf{d} & \textbf{e} & \textbf{f} & \textbf{g}\\
\midrule
\textbf{AffordGen} & $ 13/27$ & $14/27$ & $6/27$ & $\textbf{18/27}$ & $\textbf{9/27}$ & $\textbf{11/27}$ & $\textbf{16/27}$ \\
\textbf{DemoGen} & $ \textbf{14/27}$ & $0/27$ & $2/27$ & $0/27$ & $0/27$ & $0/27$ & $0/27$ \\
\textbf{CPGen} & $ 10/27$ & $2/27$ & $6/27$ & $3/27$ & $0/27$ & $2/27$ & $2/27$ \\
\textbf{FUNCTO} & $ 10/27$ & $\textbf{18/27}$ & $\textbf{8/27}$ & $9/27$ & $\textbf{9/27}$ & $5/27$ & $1/27$ \\
\bottomrule
\end{tabular}
\end{table}

\subsubsection{Mug Hanging}
\begin{figure}[H]
\begin{center}
\includegraphics[width=5cm]{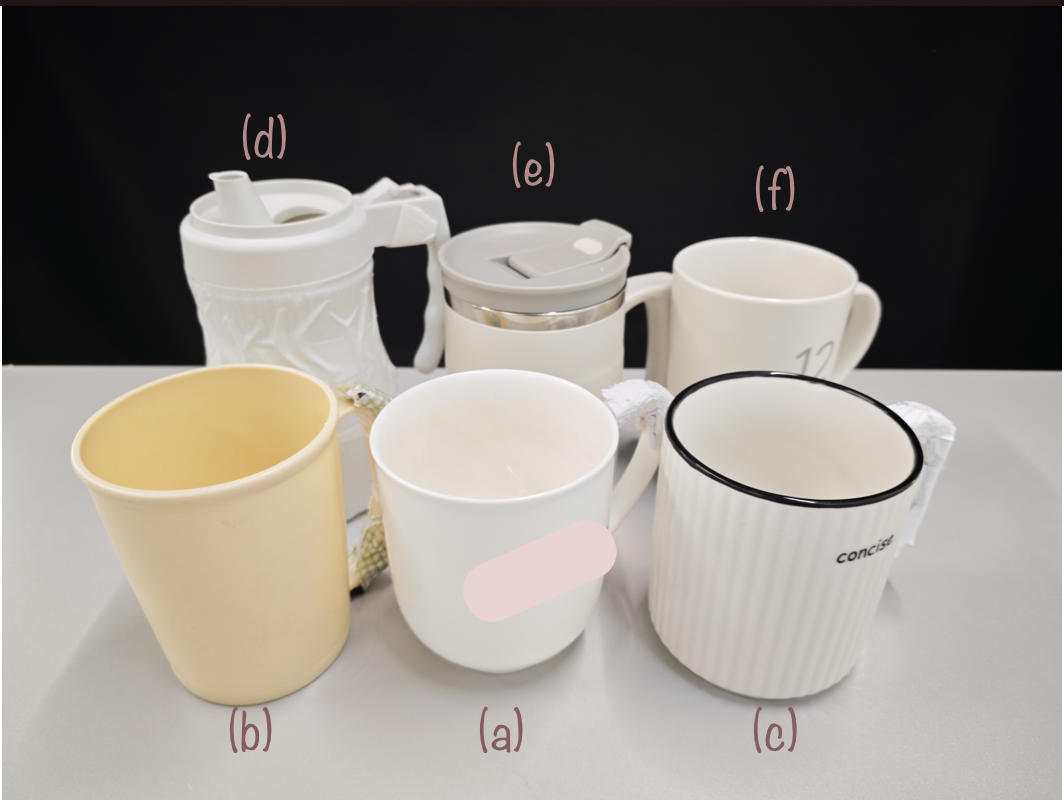}
\end{center}
\label{fig:supp_sr_mug}
\caption{Mug evaluation instances}
\end{figure}
\begin{table}[!ht]
\centering
\small 
\setlength{\tabcolsep}{3pt} 
\begin{tabular}{ccccccc}
\toprule
& \textbf{a} & \textbf{b} & \textbf{c} & \textbf{d} & \textbf{e} & \textbf{f} \\
\midrule
\textbf{AffordGen} & $ 19/27$ & $\textbf{17/27}$ & $16/27$ & $\textbf{20/27}$ & $\textbf{19/27}$ & $16/27$ \\
\textbf{DemoGen} & $ \textbf{20/27}$ & $9/27$ & $\textbf{17/27}$ & $0/27$ & $9/27$ & $\textbf{19/27}$ \\
\textbf{CPGen} & $ 19/27$ & $4/27$ & $13/27$ & $12/27$ & $5/27$ & $16/27$ \\
\textbf{FUNCTO} & $ 7/27$ & $6/27$ & $9/27$ & $10/27$ & $9/27$ & $7/27$ \\
\bottomrule
\end{tabular}
\end{table}

\subsubsection{Knife Cutting}
\begin{figure}[H]
\begin{center}
\includegraphics[width=5cm]{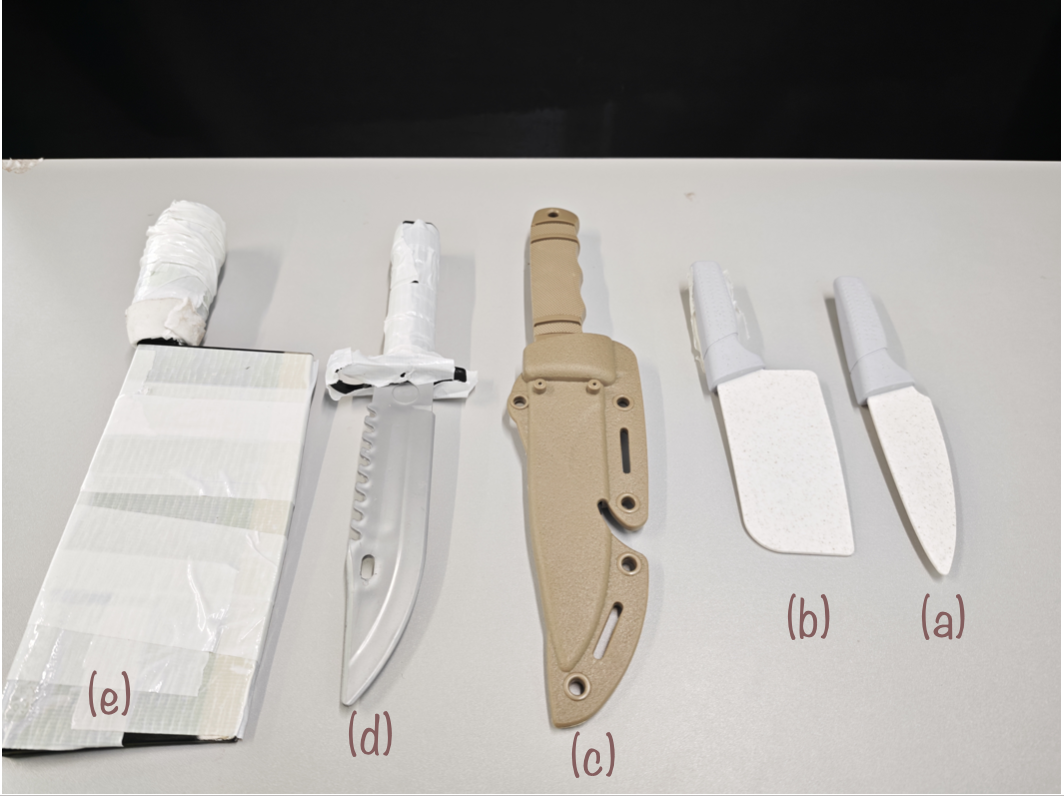}
\end{center}
\label{fig:supp_sr_knife}
\caption{Knife evaluation instances}
\end{figure}
\begin{table}[!ht]
\centering
\small 
\setlength{\tabcolsep}{4pt} 
\begin{tabular}{cccccc}
\toprule
& \textbf{a} & \textbf{b} & \textbf{c} & \textbf{d} & \textbf{e} \\
\midrule
\textbf{AffordGen} & $ 23/27$ & $\textbf{23/27}$ & $\textbf{25/27}$ & $23/27$ & $\textbf{25/27}$ \\
\textbf{DemoGen} & $\textbf{25/27}$ & $20/27$ & $10/27$ & $16/27$ & $1/27$ \\
\textbf{CPGen} & $ 23/27$ & $\textbf{23/27}$ & $24/27$ & $\textbf{24/27}$ & $17/27$ \\
\textbf{FUNCTO} & $ 21/27$ & $20/27$ & $20/27$ & $10/27$ & $11/27$ \\
\bottomrule
\end{tabular}
\end{table}

\subsubsection{Shoe Organizing}
\begin{figure}[H]
\begin{center}
\includegraphics[width=5cm]{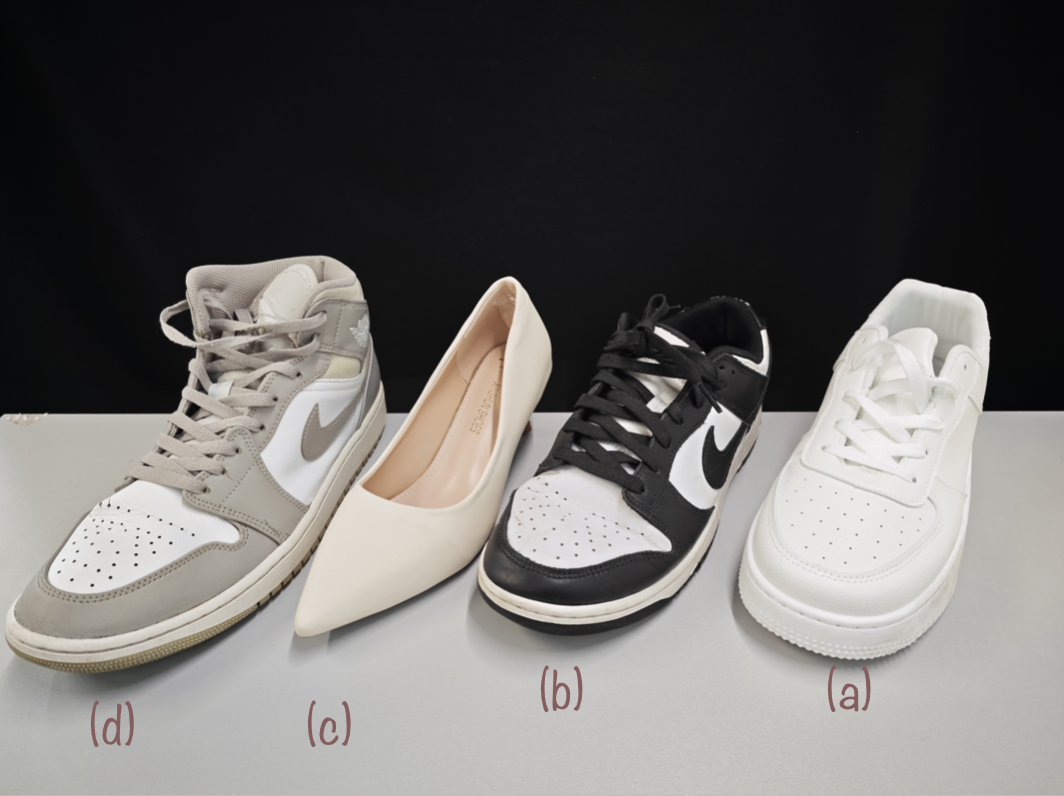}
\end{center}
\label{fig:supp_sr_shoe}
\caption{Shoe evaluation instances}
\end{figure}
\begin{table}[!ht]
\centering
\small 
\setlength{\tabcolsep}{4pt} 
\begin{tabular}{ccccc}
\toprule
 & \textbf{a} & \textbf{b} & \textbf{c} & \textbf{d} \\
\midrule
\textbf{AffordGen} & $ 11/20$ & $\textbf{14/20}$ & $\textbf{15/20}$ & $\textbf{16/20}$  \\
\textbf{DemoGen} & $13/20$ & $12/20$ & $5/20$ & $7/20$ \\
\textbf{CPGen} & $\textbf{18/20}$ & $\textbf{14/20}$ & $8/20$ & $8/20$ \\
\textbf{FUNCTO} & $15/20$ & $8/20$ & $5/20$ & $6/20$ \\
\bottomrule
\end{tabular}
\end{table}

\subsection{Baseline Implementation}
\subsubsection{DemoGen}
In both simulation and real-world experiments, we compare against the DemoGen baseline. It should be noted that the original DemoGen implementation does not generate demonstrations under varying object yaw rotations relative to the camera; the object and robot arm always face the camera from the same side. In contrast, our method extends data generation to novel camera viewpoints, and we therefore introduce a simulated rendering pipeline to produce point clouds from unseen perspectives. To ensure a fair comparison, we use the same object mesh reconstructed during the AffordGen process and generate DemoGen demonstrations across diverse positions and orientations via simulation.

\subsubsection{CPGen}
We also included the CPGen baseline in both simulation and real-world experiments. The original CPGen implementation utilizes RGB-D images and segmentation masks as policy inputs to achieve sim-to-real transfer. This requires reconstructing the real environment in simulation and providing real-time segmentation masks during real-world evaluation. To ensure a fair comparison, we converted the trajectories generated by CPGen into point cloud data for both real and simulated settings, using the same digital cousin point cloud generation method as employed in AffordGen. During the implementation of CPGen, it was necessary to apply random rescaling to the original object mesh. Since the original CPGen paper does not discuss the selection of a rescaling factor, we set CPGen's rescaling factor such that the bounding box of the original mesh could cover the bounding box of the test meshes.

\subsubsection{FUNCTO}
We included the FUNCTO baseline in our real-world experiments. The original FUNCTO baseline, as an online planning algorithm, incorporates a complex keypoint selection pipeline that involves fine-grained adjustments of planning angle alignment using large language models. To ensure a fair comparison, we simplified FUNCTO's keypoint selection process by directly employing DINOv2 for keypoint correspondence, which aligns with AffordGen's approach of selecting keypoints from 2D camera views. It should be noted that the AffordGen pipeline does not inherently constrain the methodology for semantic keypoint selection; the sophisticated selection process used in the original FUNCTO implementation, albeit at the cost of annotation efficiency, is equally applicable to our AffordGen data generation pipeline.

\subsection{Real World Experiment Demos}
Please refer to the associated videos to see the performance of policies trained on AffordGen-generated data in the real world. With a small number of expert demonstrations on a single object, the data generated by AffordGen can train policies that perform well both on objects within the same category and across different categories. The learned policy is able to execute consecutively on different objects, showcasing the superior generalization ability of \ours.

\section{3D Mesh Dataset}
\label{app:dataset}
\subsection{3D Mesh Dataset Pre-processing}
To obtain a sufficient number of meshes for a specific category, we leveraged an existing 3D generative model ~\cite{hunyuan3d22025tencent}. For the teapot and mug categories, the generated meshes are almost upright, with their in-plane rotations (within the XY plane) typically aligning with one of the four cardinal angles: 0°, 90°, 180°, or 270°. To estimate their poses, we employ the off-the-shelf 6D pose estimator Omni6DPose~\cite{zhang2024omni6dpose} by rendering images from predefined viewpoints around the object. The estimated rotation is then applied in reverse to align the mesh to a canonical pose. For the knife category, the generated meshes may have different orientations in $SE(3)$, which hinders the accuracy of pose canonicalization. Thus, Principal Component Analysis (PCA) is first used to extract the three principal axes of each generated knife, they are then rotated to match the axes of the reference mesh. Since the curvature of the blade in the longest and middle axes tends to introduce estimation errors, we align the models first along the shortest axis, which provides the most accurate estimate due to its shape consistency.

\subsection{Dataset Visualization}
\begin{figure*}[htbp]
    \centering
    \includegraphics[width=0.8\textwidth]{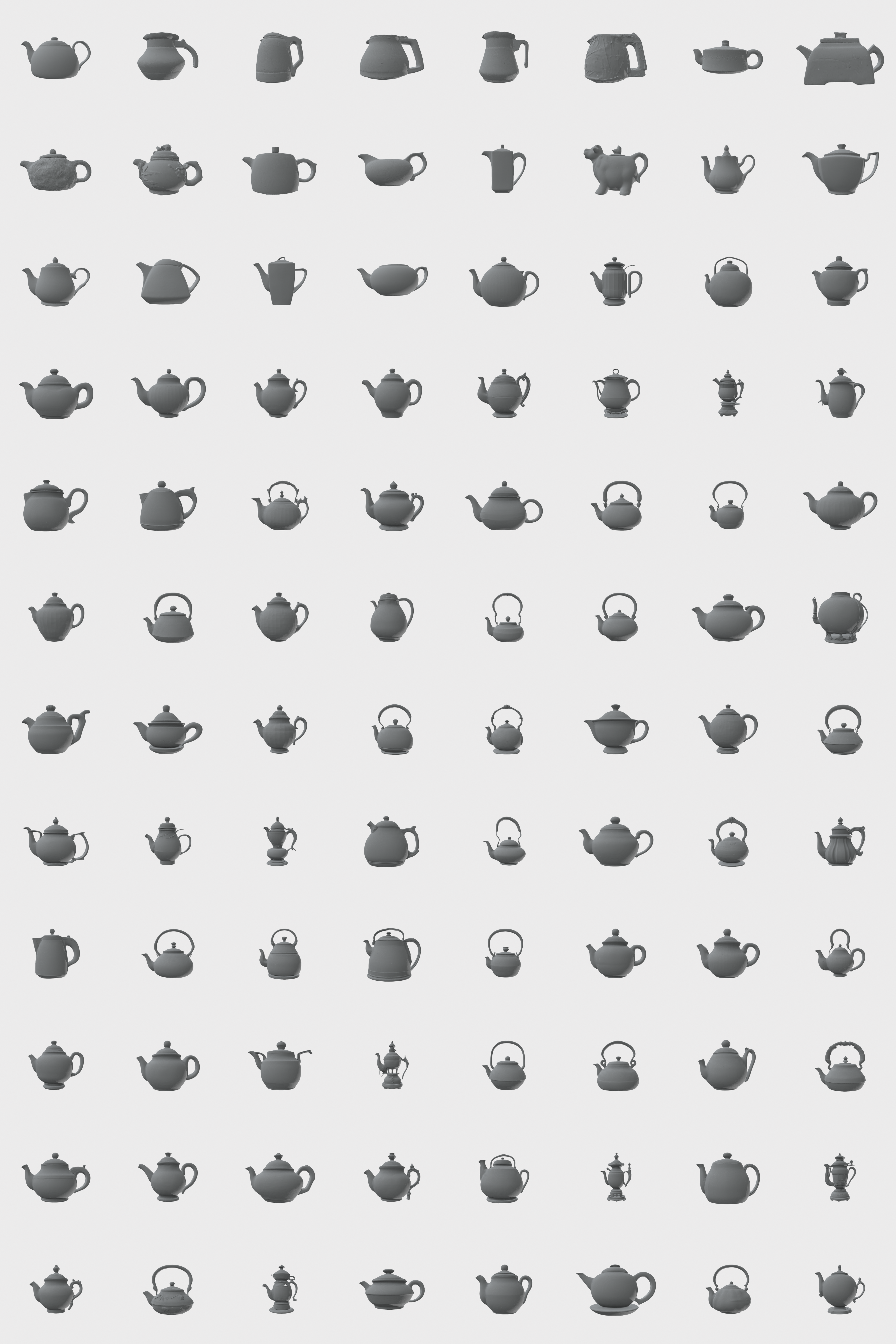}
    \caption{Part of the teapot meshes used for demonstration generation}
    \label{fig:teapot_mesh_grid}
\end{figure*}

We present the mesh dataset used for generating demonstrations for the teapot task in Figure~\ref{fig:teapot_mesh_grid}.

\subsection{Correspondence Prediction Visualization}
\begin{figure}[H]
\begin{center}
\includegraphics[width=8cm]{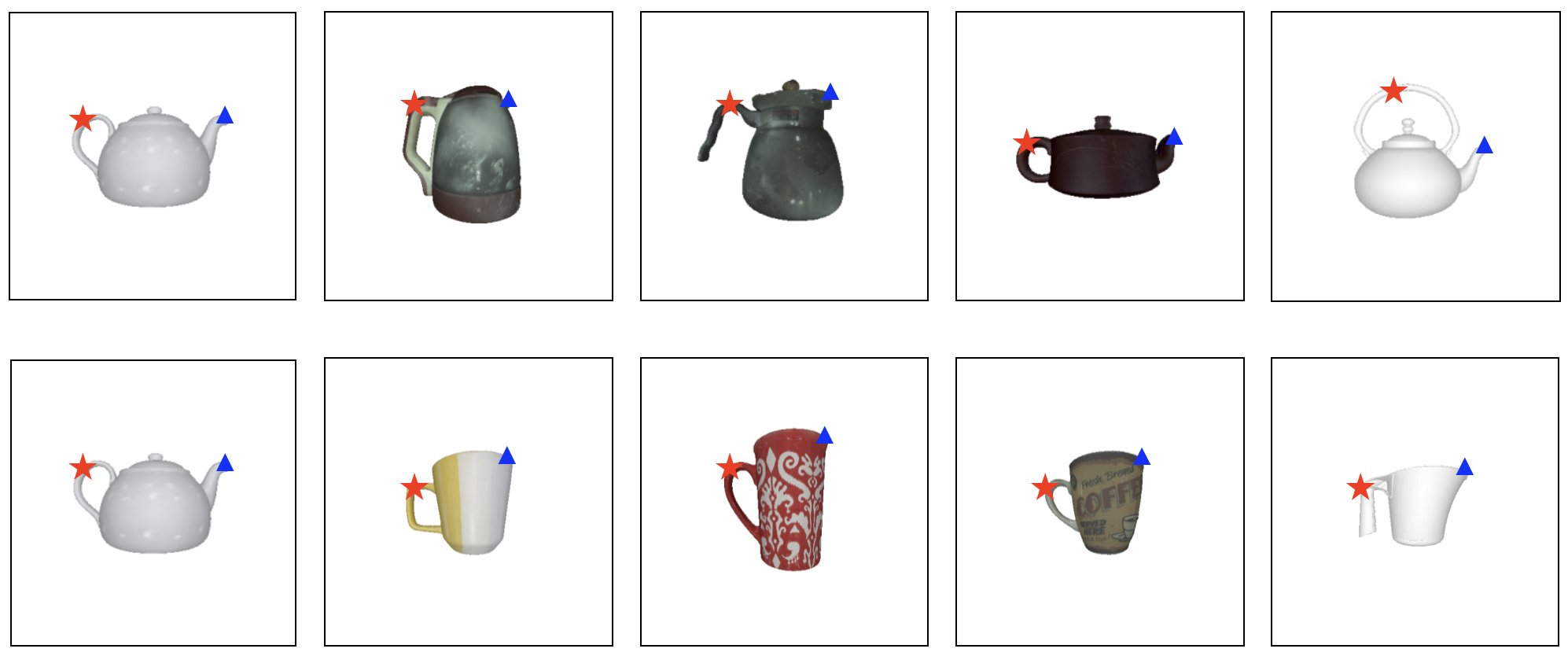}
\end{center}
\label{fig:supp_teapot_vis}
\end{figure}
\begin{figure}[H]
\begin{center}
\includegraphics[width=8cm]{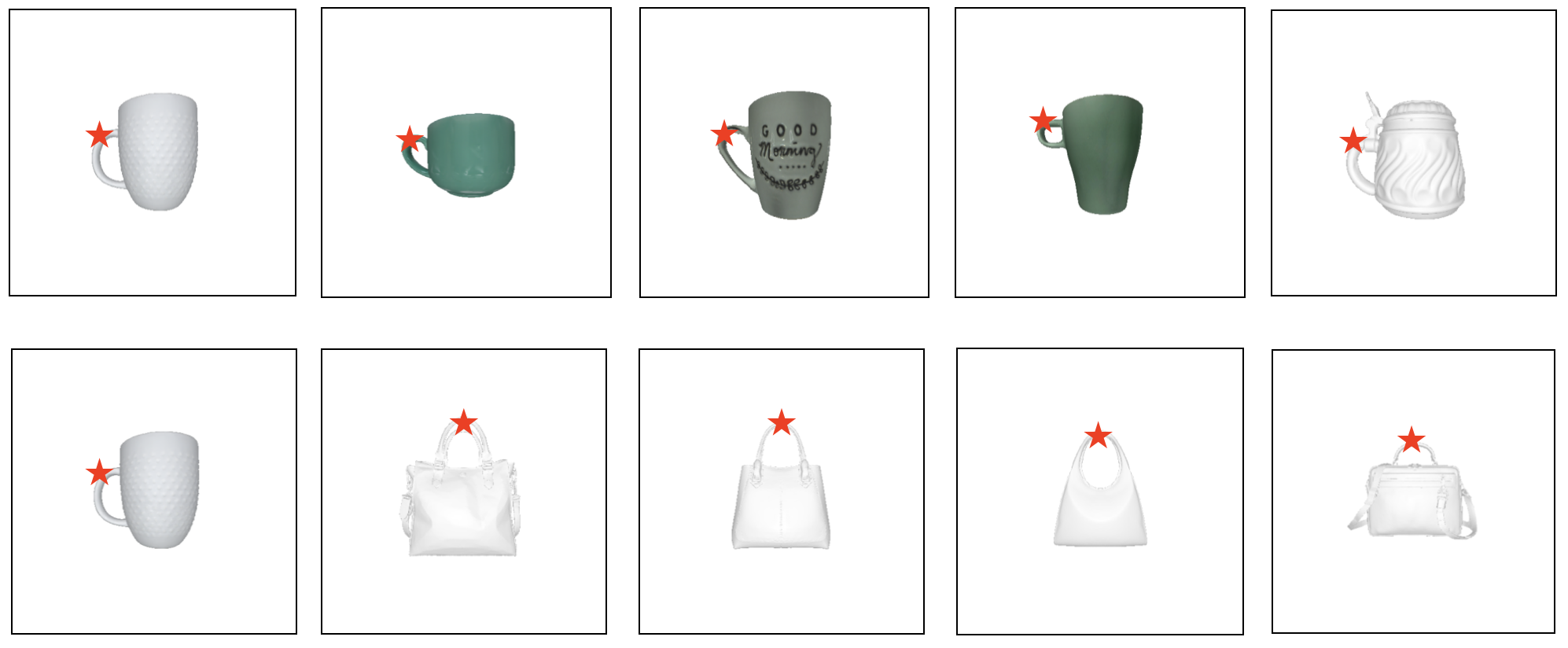}
\end{center}
\label{fig:supp_mug_vis}
\end{figure}
\begin{figure}[H]
\begin{center}
\includegraphics[width=8cm]{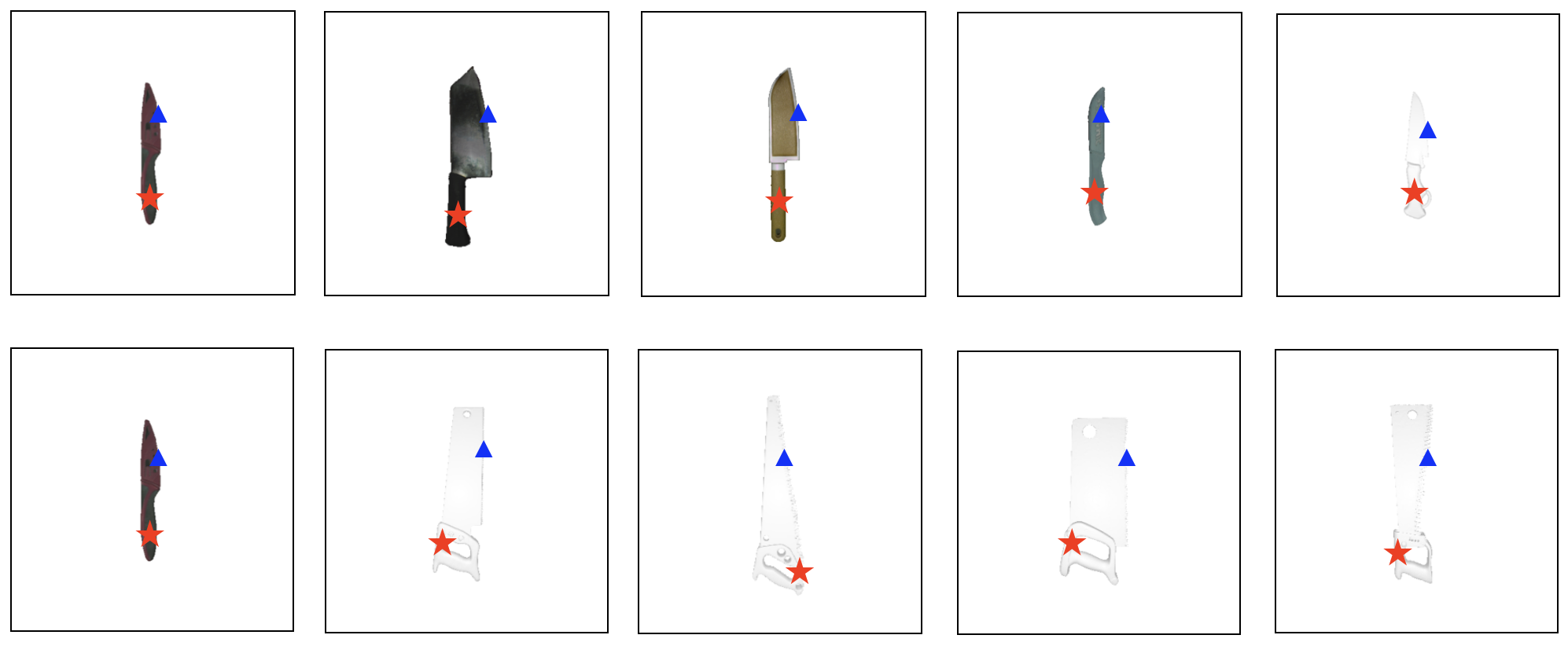}
\end{center}
\label{fig:supp_knife_vis}
\end{figure}
\begin{figure}[H]
\begin{center}
\includegraphics[width=8cm]{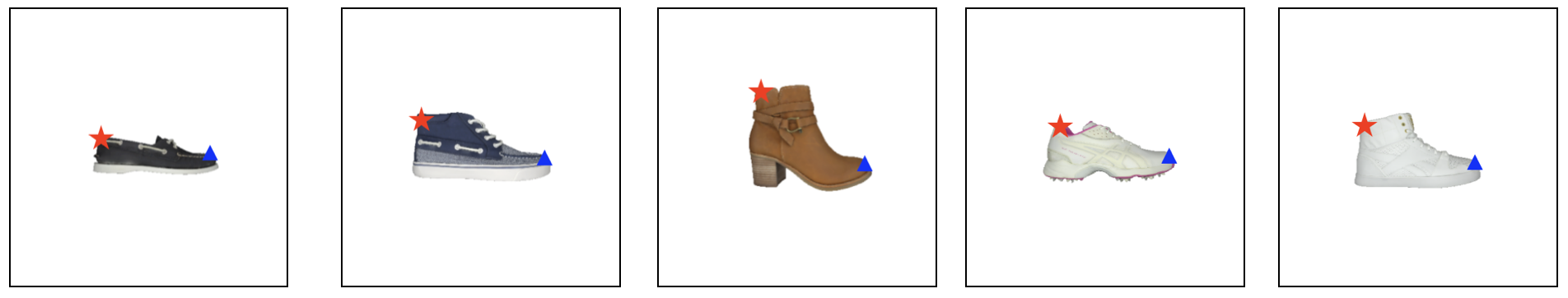}
\end{center}
\label{fig:supp_shoe_vis}
\end{figure}

\section{Data Generation Details}
\label{app:data_generation}

\begin{figure*}[ht]
\begin{center}
\includegraphics[width=\linewidth]{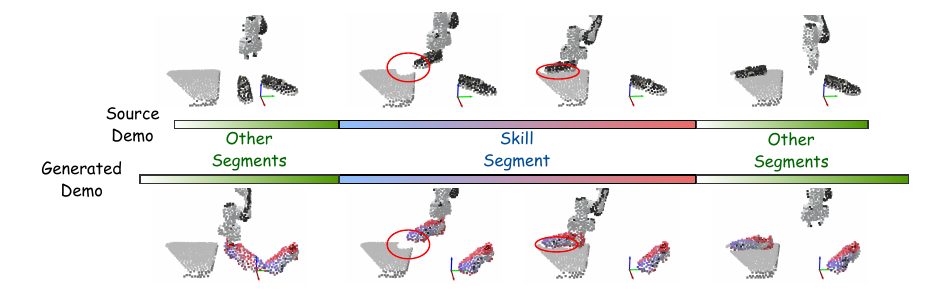}
\end{center}
\caption{We preserve the occlusions of the goal object during skill segment in our point cloud generation process.}
\label{fig:supp_pcd_gen}
\end{figure*}

Given that the generated trajectories may vary in length from the original ones, we sample the goal object point cloud from random timestamps (excluding the skill segment) of the source demonstration to serve as the goal object point cloud for the new demonstration. To preserve the visual authenticity of the occlusions that occur when the object interacts with the goal object during the skill segment, the goal object point cloud in this phase is sequentially replayed from the corresponding segment of the source demonstration. The overall process can be expressed as follows:

\[
\begin{cases}
 \operatorname{Assemble}\left( \mathcal{O}_{\text{real}}[i], \ \mathcal{O}_{\text{sim}}[t] \right), \quad \text{if } t \notin \mathcal{T}_s' \\
 \operatorname{Assemble}\left( \mathcal{O}_{\text{real}}[t - t_s' + t_s], \ \mathcal{O}_{\text{sim}}[t] \right), & \text{if } t \in \mathcal{T}_s'
\end{cases}
\]
where $t_s$ and $t_s'$ are the start timesteps of $\tau_s$ and $\tau_s'$, respectively, and $i \in \mathcal{U}(0, T')$ denotes uniform sampling. The $\operatorname{Assemble}$ operator represents the process of concatenating the point cloud of the robot arm and the new meshes (from simulation) with the goal object point cloud (from the real world), under the constraint that the point count per frame remains consistent with the source demonstration. A visualization of the point cloud generation strategy is presented in Figure~\ref{fig:supp_pcd_gen}.

\newpage


\end{document}